%% file: sample-sigconf.tex
\documentclass[sigconf]{acmart}
\input{0__package.tex}

\AtBeginDocument{%
  \providecommand\BibTeX{{%
    \normalfont B\kern-0.5em{\scshape i\kern-0.25em b}\kern-0.8em\TeX}}}

\copyrightyear{2023}
\acmYear{2023}
\setcopyright{acmlicensed}
\acmConference[KDD '23] {Proceedings of the 29th ACM SIGKDD Conference on Knowledge Discovery and Data Mining}{August 6--10, 2023}{Long Beach, CA, USA.}
\acmBooktitle{Proceedings of the 29th ACM SIGKDD Conference on Knowledge Discovery and Data Mining (KDD '23), August 6--10, 2023, Long Beach, CA, USA}
\acmPrice{15.00}
\acmISBN{979-8-4007-0103-0/23/08}
\acmDOI{10.1145/3580305.3599410}

\begin{document}

\title{Learning Strong Graph Neural Networks with Weak Information}


\author{Yixin Liu}
\affiliation{%
  \institution{Monash University}
  \country{}
}
\email{yixin.liu@monash.edu}

\author{Kaize Ding}
\affiliation{%
  \institution{Arizona State University}
  \country{}
}
\email{kaize.ding@asu.edu}

\author{Jianling Wang}
\affiliation{%
  \institution{Texas A\&M University}
  \country{}
}
\email{jlwang@tamu.edu}

\author{Vincent Lee}
\affiliation{%
  \institution{Monash University}
  \country{}
}
\email{vincent.cs.lee@monash.edu}

\author{Huan Liu}
\affiliation{%
  \institution{Arizona State University}
  \country{}
}
\email{huanliu@asu.edu}

\author{Shirui Pan*}
\affiliation{%
  \institution{Griffith University}
  \country{}
}
\email{s.pan@griffith.edu.au}

\makeatletter
\def\authornotetext#1{
	\g@addto@macro\@authornotes{%
	\stepcounter{footnote}\footnotetext{#1}}%
}
\makeatother

\authornotetext{Shirui Pan is the corresponding author.}
\renewcommand{\shortauthors}{Liu et al.}
\renewcommand{\authors}{Yixin Liu, Kaize Ding, Jianling Wang, Vincent Lee, Huan Liu, Shirui Pan}

\begin{abstract}
\input{0_abs.tex}
\end{abstract}





\maketitle

\section{Introduction}
\input{1_intro.tex}

\section{Related Works}
\input{2_rw.tex}

\section{Preliminaries}
\input{3_pre.tex}

\section{Design Motivation and Analysis} \label{sec:base}
\input{4_moti.tex}

\section{Methodology}
\input{5_method.tex}

\section{Experiments}
\input{6_exp.tex}

\section{Conclusion}
\input{7_conclu.tex}

\vspace{-2mm}
\begin{acks}
This work is supported by ARC Future Fellowship (No. FT210100097), Amazon Research Award, NSF (No. 2229461), and ONR (No. N00014-21-1-4002).
\end{acks}

\bibliographystyle{ACM-Reference-Format}
\balance
\bibliography{ref_simp}

\appendix

\input{8_appendix.tex}

\end{document}

%% file: 0__package.tex
\usepackage{subfigure}
\usepackage{color}
\usepackage{xspace}
\usepackage{makecell}
\usepackage{multirow}
\usepackage{tcolorbox}
\usepackage{threeparttable} 
\usepackage[linesnumbered,algoruled,boxed,lined,ruled]{algorithm2e}
\SetKwInOut{Parameter}{Parameters}

\SetCommentSty{mycommfont}

\newcommand{\tabincell}[2]{\begin{tabular}{@{}#1@{}}#2\end{tabular}}  

\newcommand{\ourmethod}{D$^2$PT\xspace}
\newcommand{\ourbase}{DPT\xspace}

\definecolor{sotacolor}{RGB}{152,52,48}
\definecolor{runupcolor}{RGB}{236,84,40}
\newcommand{\sota}[1]{\textcolor{sotacolor}{\mathbf{#1}}}
\newcommand{\runup}[1]{\textcolor{runupcolor}{\underline{#1}}}

%% file: 0_abs.tex
Graph Neural Networks (GNNs) have exhibited impressive performance in many graph learning tasks. 
Nevertheless, the performance of GNNs can deteriorate when the input graph data suffer from weak information, i.e., incomplete structure, incomplete features, and insufficient labels. 
Most prior studies, which attempt to learn from the graph data with a specific type of weak information, are far from effective in dealing with the scenario where diverse data deficiencies exist and mutually affect each other. 
To fill the gap, in this paper, we aim to develop an effective and principled approach to the problem of graph learning with weak information (GLWI). 
Based on the findings from our empirical analysis, we derive two design focal points for solving the problem of GLWI, i.e., enabling long-range propagation in GNNs and 
allowing information propagation to those stray nodes isolated from the largest connected component. Accordingly, we propose \ourmethod, a dual-channel GNN framework that performs long-range information propagation not only on the input graph with incomplete structure, but also on a global graph that encodes global semantic similarities. We further develop a prototype contrastive alignment algorithm that aligns the class-level prototypes learned from two channels, such that the two different information propagation processes can mutually benefit from each other and the finally learned model can well handle the GLWI problem. Extensive experiments on eight real-world benchmark datasets demonstrate the effectiveness and efficiency of our proposed methods in various GLWI scenarios. 

%% file: 1_intro.tex
\begin{figure}[t]
\vspace{-2mm}
	\centering
	\includegraphics[width=0.9\columnwidth]{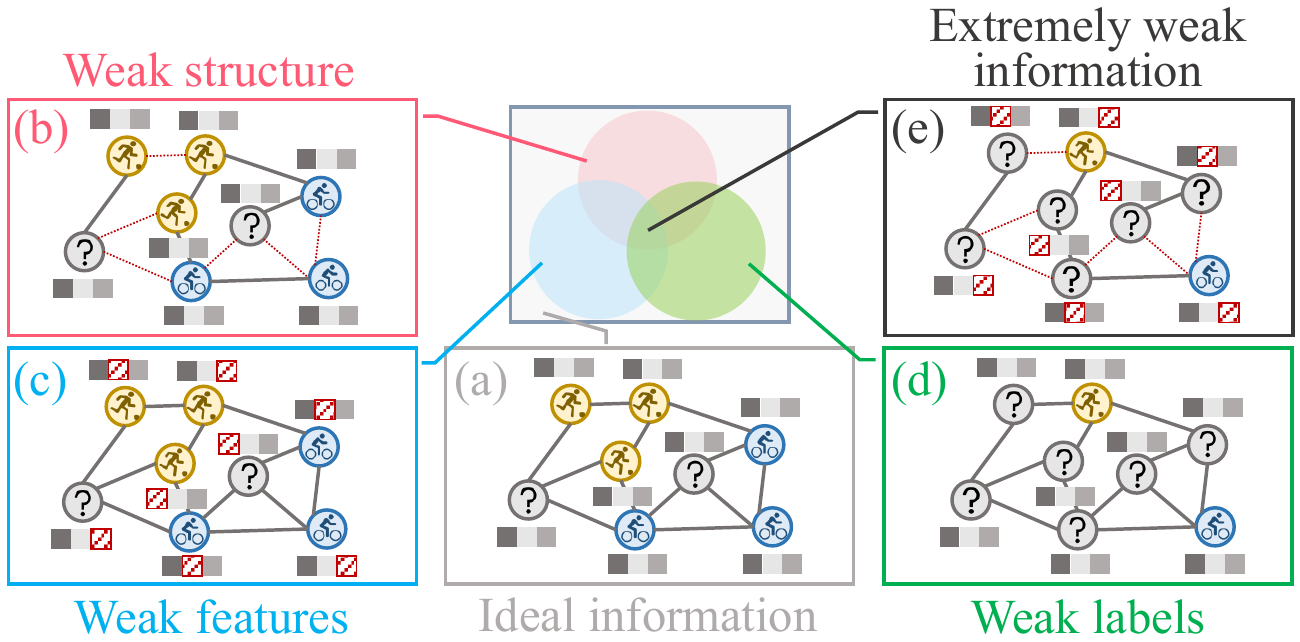}
 \vspace{-0.3cm}
	\caption{Sketch maps of graph data with (a) ideal information, (b) weak structure, (c) weak features, (d) weak labels, and (e) extremely weak information.}
	\label{fig:sketch}
    \vspace{-0.5cm}
\end{figure}

Graph neural networks (GNNs) have attracted increasing research attention in recent years~\cite{wu2021comprehensive, kipf2017semi, xu2019how}. Attributed to the message propagation and feature transformation mechanisms, GNNs are capable to learn informative representations for graph-structured data and tackle various graph learning tasks such as node classification~\cite{kipf2017semi, velivckovic2018graph}, link prediction~\cite{zhang2018link, luo2023graph}, and graph classification~\cite{xu2019how, tan2022federated}. GNNs have shown remarkable success across diverse knowledge discovery and data mining scenarios, including drug discovery~\cite{gaudelet2021utilizing}, fraud detection~\cite{wang2019semi, liu2023good}, and knowledge reasoning~\cite{luo2023npfkgc}.

The majority of GNNs rely on a fundamental assumption, i.e., the observed graph data are ideal enough to provide \textit{sufficient information} ({including structure, features, and labels}) for model training (Fig.~\ref{fig:sketch}(a))~\cite{wang2021graph}. 
Such an assumption, unfortunately, is often invalid in practical scenarios, because real-world graph data extracted from complex systems usually contain incomplete and insufficient information~\cite{marsden1990network, you2020handling, li2019label,ding2022data,ding2022toward}. 
{For instance, due to privacy concerns or human errors in data collection process, the edges and node features in real-world graphs are sometimes missing, leading to incomplete graph data with \textit{weak structure} (Fig.~\ref{fig:sketch}(b)) and \textit{weak features} (Fig.~\ref{fig:sketch}(c)).}
Besides, the label information for model training is not always sufficient, especially in domains where data annotation is costly (e.g., chemistry and biology~\cite{zitnik2018prioritizing}). The deficiency of labels gives rise to graph data with \textit{weak labels} (Fig.~\ref{fig:sketch}(d)). 
The incompleteness and insufficiency, inevitably, hinder us from training expressive GNN models on these graph data with \textit{weak information}~\cite{jin2020graph, sun2020multi}. \looseness-2

To learn strong graph learning models with such weak information, several recent efforts propose to equip GNNs with diverse specialized designs to handle data deficiency. 
Some existing methods introduce supplemental designs that learn to recover the missing information, such as constructing new edges with graph structure learning~\cite{jin2020graph, chen2020iterative}, imputing missing features with attribute completion~\cite{chen2020learning, spinelli2020missing, taguchi2021graph}, and enriching training set with pseudo labeling~\cite{sun2020multi, ding2022meta}. Another branch of approaches aims to fully leverage the observed weak information with techniques like Poisson learning for few-label learning~\cite{wan2021contrastive} and distillation for knowledge mining on features and structure~\cite{huo2022t2}. 
Nevertheless, most of them solely consider data deficiency in a single aspect, despite the fact that deficient structure, features, and labels can occur simultaneously in real-world scenarios. 
{To bridge the gap, a natural research question is: ``Can we design a universal and effective GNN for graph learning with weak information (GLWI)?''}

To answer this question, in this paper, we first conduct a comprehensive analysis to investigate the performance of GNNs {when learning with weak information}. 
With empirical discussion, we find information propagation, the fundamental operation in GNNs, plays a crucial role in {mitigating the data incompleteness}. 
However, the limitations of model architectures and deficiencies of data hinder conventional GNNs to execute effective information propagation on incomplete data. 
{From the perspective of model architectures, we pinpoint that GNNs with \textit{long-range propagation} enable sufficient information communication, which not only helps recover missing data but also further exploits the observed information.} 
From the perspective of data, we ascribe the performance degradation of graph learning with extremely weak information (Fig.~\ref{fig:sketch}(e)) to the incomplete graph structure. 
Concretely, the scattered nodes isolated from the largest connected component (i.e., \textit{stray nodes}) lead to ineffective information propagation, which impedes feature imputation and supervision signal spreading. 
These empirical findings shed light on the key criteria that help improve GNNs to handle the GLWI problem. 

Following the above design principles, we propose a powerful yet efficient GNN model, \textit{\textbf{D}ual-channel \textbf{D}iffused \textbf{P}ropagation then \textbf{T}ransformation} (\ourmethod for short), for GLWI. 
Our theme is to enable effective information propagation on graph data with weak information by conducting efficient long-range propagation and relieving the stray node problem. 
More specifically, to enhance the expressive capability and reduce the computational cost of long-range message passing, we design a graph diffusion-based backbone model termed \ourbase which enables effective message passing while preserving high running efficiency. 
{To allow information propagation on stray nodes, based on propagated features, we further learns a global graph by connecting nodes sharing similar semantics from a global view. }
We apply dual-channel training and contrastive prototype alignment mechanisms to \ourmethod, which fully leverages the knowledge of the global graph to optimize the \ourbase backbone. 
Extensive experiments on 8 real-world benchmark datasets demonstrate the effectiveness, generalization capability, and efficiency of \ourmethod.

To summarize, our paper makes the following contributions:

\begin{itemize}
    \item \textbf{Problem.} {We make the first attempt to investigate the graph learning problem with extremely weak information where structure, features, and labels are incomplete simultaneously, advancing existing research scope from a single angle to multiple intertwined perspectives as a whole.}
    \item \textbf{Analysis.} We provide a comprehensive analysis to investigate the impact of data deficiency on GNNs, which further guides our algorithm designs against GLWI problem. 
    \item \textbf{Algorithms.} We propose a novel method termed \ourmethod, which provides a universal, effective, and efficient solution for diverse GLWI scenarios. 
    \item \textbf{Experiments.} We conduct extensive experiments to demonstrate that \ourmethod can offer superior performance over baseline methods in multiple GLWI tasks.
\end{itemize}

%% file: 2_rw.tex
\subsection{Graph Neural Networks}
Graph neural networks (GNNs) are a family of neural networks that learn complex dependencies in graph-structured data~\cite{kipf2017semi, velivckovic2018graph, wu2021comprehensive, zhang2022trustworthy}. Based on the message passing paradigm, existing GNNs are composed of two types of atomic operations: propagation (\texttt{P}) that aggregates representations to adjacent nodes and transformation (\texttt{T}) that updates node representations with learnable non-linear mappings~\cite{xu2019how, zhang2022deep, zhang2022model}. Different GNNs have their specific designs of \texttt{P}/\texttt{T} functions and orders of \texttt{P}/\texttt{T} operations~\cite{wu2021comprehensive}. Commonly used \texttt{P} functions include averaging~\cite{kipf2017semi}, summation~\cite{xu2019how}, and attention~\cite{velivckovic2018graph}, while \texttt{T} is often defined as perceptron layer(s)~\cite{hamilton2017inductive,wu2019simplifying}. To organize \texttt{P}/\texttt{T} operations, the majority of GNNs follow a \texttt{PTPT} scheme, where multiple entangled ``\texttt{P}-\texttt{T}'' layers are sequentially stacked ~\cite{kipf2017semi, velivckovic2018graph, xu2019how, hamilton2017inductive}. There are also some GNNs that execute multi-round \texttt{P}/\texttt{T} operation first, and execute another type of operation in the following step, i.e., \texttt{PPTT} scheme~\cite{wu2019simplifying, zhu2020simple} and \texttt{TTPP} scheme~\cite{gasteiger2018predict, chien2020adaptive}. 
Recently efforts extend GNNs to various learning scenarios, such as unsupervised representation learning~\cite{zheng2022unifying, zheng2022rethinking}, adversarial attack~\cite{zhang2021projective, zhang2022projective}, and architecture search~\cite{zheng2023auto, zheng2022multi}.

\subsection{Graph Learning with Weak Information }

Graph learning with weak information (GLWI) aims to learn graph machine learning models when input graph data with 1) incomplete structure, 2) incomplete features, and/or 3) insufficient labels. Most existing works focus on learning GNNs on graphs with data insufficiency in a single aspect. 

To handle \textit{incomplete structure}, \textbf{graph structure learning} aims to jointly learn an optimized graph structure along with the backbone GNN~\cite{zhu2021deep, liu2022towards}. As representative methods, LDS~\cite{franceschi2019learning} and GEN~\cite{wang2021graph} use Bernoulli model and stochastic block model respectively to parameterize the adjacency matrix, and train the probabilistic models along with the backbone GNNs. IDGL~\cite{chen2020iterative} and Simp-GCN~\cite{jin2021node} introduce metric learning technique to revise the original graph structure. Pro-GNN~\cite{jin2020graph} directly models the adjacency matrix with learnable parameters and learns it with GNN alternatively. 

To resolve \textit{incomplete features}, \textbf{attribute completion} aims to recover the missing data from the existing ones~\cite{chen2020learning,ding2022data}. Spinelli et al.~\cite{spinelli2020missing} first apply a GNN-based autoencoder for missing data imputation. SAT~\cite{chen2020learning} introduces a feature-structure distribution matching mechanism to the node attribute completion model. GCN$_{MF}$~\cite{taguchi2021graph} uses Gaussian Mixture Model to transform the incomplete features at the first layer of GNN. HGNN-AC~\cite{jin2021heterogeneous} employs topological embeddings to benefit attribute completion. 

A line of studies termed \textbf{label-efficient graph learning} propose to learn GNN models from data with \textit{insufficient labels}~\cite{li2019label, sun2020multi,ding2022meta,ding2022robust}. IGCN~\cite{li2019label} is a pioneering work that applies a label-aware low-pass graph filter on GNNs to achieve label efficiency. M3S~\cite{sun2020multi} leverages clustering technique to provide extra supervision signals and trains the model in a multi-stage manner. CGPN~\cite{wan2021contrastive} utilizes Poisson network and contrastive learning for label-efficient graph learning. Meta-PN~\cite{ding2022meta} generates high-quality pseudo labels with label propagation strategy to augment the scarce training samples. 

Despite their success in handling GLWI from a single aspect, to the best of our knowledge, none of the existing works has jointly considered the data insufficiency from three aspects. Moreover, with carefully-crafted learning procedures, most of them require high computational costs for training, damaging their running efficiency on large-scale graphs. To bridge the gaps, in this paper, we aim to propose a general, efficient, and effective approach for GLWI. 

%% file: 3_pre.tex
\noindent\textbf{Notations}. 
We consider an attributed and undirected graph as $\mathcal{G} = (\mathcal{V},\mathcal{E},\mathbf{X}) = (\mathbf{A},\mathbf{X})$, where $\mathcal{V}=\{v_1,\cdots,v_n\}$ is the node set with size $n$, $\mathcal{E}$ is the edge set with size $m$, $\mathbf{A} \in \{0, 1\}^{n \times n}$ is the binary adjacency matrix (where the $i,j$-th entry $\mathbf{A}_{ij}=1$ means $v_i$ and $v_j$ are connected and vice versa), and $\mathbf{X} \in \mathbb{R}^{n \times d}$ is the feature matrix (where the $i$-th row $\mathbf{X}_i$ is the $d$-dimensional feature vector of node $v_i$). The label of $\mathcal{G}$ is represented by a label matrix $\mathbf{Y} \in \mathbb{R}^{n \times c}$, where $c$ is the number of classes, each row is a one-hot vector, and the $i,j$-th entry $\mathbf{Y}_{ij}=1$ indicates that node $v_i$ belongs to the $j$-th class and vice versa. The neighbor set of node $v_i$ is represented by $\mathcal{N}_{v_i} = \{v_j|\mathbf{A}_{ij}=1\}$. The normalized adjacency matrix is represented by $\tilde{\mathbf{A}} = \mathbf{D}^{-1/2}\mathbf{A}\mathbf{D}^{-1/2}$, where $\mathbf{D}$ is the diagonal degree matrix $\mathbf{D}_{ii}=\sum_{j} \mathbf{A}_{ij}$.

\noindent\textbf{Graph neural networks (GNNs)}. 
Following the message passing paradigm, GNNs can be defined as the stacked combination of two fundamental operations: \textbf{propagation} (\texttt{P}) that aggregates the representations of each node to its neighboring nodes and \textbf{transformation} (\texttt{T}) that transforms the node representations with non-linear mappings~\cite{zhang2022deep}. With $\mathbf{h}_i^{(i)}$ and $\mathbf{h}_i^{(o)}$ as the input and output representations of node $v_i$ respectively, the \texttt{P} operation can be formulated by $\mathbf{h}_i^{(o)} \gets \texttt{P} (\mathbf{h}_i^{(i)}, \{\mathbf{h}_j^{(i)}| v_j \in \mathcal{N}_{v_i} \})$, and the  \texttt{T} operation can be formulated by $\mathbf{h}_i^{(o)} \gets \texttt{T} (\mathbf{h}_i^{(i)})$. Taking GCN~\cite{kipf2017semi} as an implementation, the \texttt{P} and \texttt{T} can be written as $\mathbf{h}_i^{(o)} = \sum_{j} \tilde{\mathbf{A}}_{ij}\mathbf{h}_j^{(i)} $ and $\mathbf{h}_i^{(o)} = \sigma (\mathbf{W} \mathbf{h}_i^{(i)})$, where $\mathbf{W}$ is a learnable parameter matrix and $\sigma(\cdot)$ is a non-linear activation function.  

According to the manner of stacking \texttt{P} and \texttt{T} operations, GNNs can be divided into two categories: entangled GNNs (\texttt{PTPT}) and disentangled GNNs (\texttt{PPTT} or \texttt{TTPP})~\cite{zhang2022model}. For example, a two-layer GCN~\cite{kipf2017semi} is an entangled GNN that can be written as $\mathbf{H}^{(o)} = \texttt{GCN}(\mathbf{H}^{(i)}) = \texttt{P}(\texttt{T}(\texttt{P}(\texttt{T}(\mathbf{H}^{(i)}))))$, where $\texttt{P}$ and $\texttt{T}$ are stacked alternatively and in couples. A two-layer SGC~\cite{wu2019simplifying} is a disentangled \texttt{PPTT} GNN that can be written as $\mathbf{H}^{(o)} = \texttt{SGC}(\mathbf{H}^{(i)}) = \texttt{T}(\texttt{P}(\texttt{P}(\mathbf{H}^{(i)})))$, where $\mathtt{T}$ is executed after all $\mathtt{P}$ are finished. Given a GNN, we define the iteration times of $\texttt{P}$ and $\texttt{T}$ as its \textbf{propagation step} $s_p$ and \textbf{transformation step} $s_t$, respectively.

\noindent\textbf{Semi-supervised node classification}. 
In this paper, we focus on the semi-supervised node classification task, which is an essential and widespread task in graph machine learning~\cite{kipf2017semi, hamilton2017inductive, velivckovic2018graph, wu2019simplifying, franceschi2019learning, sun2020multi}. In this task, only the labels of a small fraction of nodes $\mathcal{V}_{L} \subset \mathcal{V}$ are available for model training, and the goal in inference phase is to predict the labels of unlabeled nodes $\mathcal{V}_{U} \subset \mathcal{V}$, w.r.t. $\mathcal{V}_{U} \cap \mathcal{V}_{L} = \emptyset$. We denote the training labels as $\mathbf{Y}_L \in \mathbb{R}^{n_L \times c}$ where $n_L=|\mathcal{V}_{L}|$.

\noindent\textbf{Graph learning with weak information (GLWI)}. 
To formulate GLWI, we first define ideal graph data for semi-supervised node classification under some ideal conditions. 

\begin{definition}[Ideal graph data]
Let ideal graph data be $\hat{\mathcal{D}} = (\hat{\mathcal{G}}, \hat{\mathbf{Y}}_L) = (({\mathcal{V}}, \hat{\mathcal{E}}, \hat{\mathbf{X}}), \hat{\mathbf{Y}}_L)$, where $\hat{\mathcal{E}}$ is an ideal edge set that contains all necessary links, $\hat{\mathbf{X}}$ is an ideal feature matrix that contains all informative features, and $\hat{\mathbf{Y}}_L$ is an ideal label matrix that contains adequate labels (with number $\hat{n}_L$) with a balanced distribution. 
\end{definition}

Note that ideal graph data is a perfect case for graph learning. In real-world scenarios, the data for model training (i.e., observed graph data) are sometimes incomplete and insufficient. 
Specifically, the structure can be incomplete in graph data with an \textit{incomplete edge set} $\check{\mathcal{E}} \in \hat{\mathcal{E}}$ that contains limited edges to provide adequate information for graph learning. 
Meanwhile, some critical elements in the feature matrix are missing, which can be represented by an \textit{incomplete feature matrix} $\check{\mathbf{X}} = \mathbf{M} \odot \hat{\mathbf{X}}$, where $\mathbf{M} \in \{0, 1\}^{n \times d}$ is the missing mask matrix. 
Besides, the available labels for model training can be scarce, indicating an \textit{insufficient label matrix} $\check{\mathbf{Y}}_L$ with training number $\check{n}_L \ll \hat{n}_L$. 
Based on the above definitions, the basic GLWI scenarios can be formulated by:

\begin{definition}[Basic GLWI scenarios] \label{def:basic}
Let graph data with weak structure, weak features, and weak labels be $\mathcal{D}_{ws}=(({\mathcal{V}}, \check{\mathcal{E}}, \hat{\mathbf{X}}), \hat{\mathbf{Y}}_L)$, $\mathcal{D}_{wf}=(({\mathcal{V}}, \hat{\mathcal{E}}, \check{\mathbf{X}}), \hat{\mathbf{Y}}_L)$, and $\mathcal{D}_{wl}=(({\mathcal{V}}, \hat{\mathcal{E}}, \hat{\mathbf{X}}), \check{\mathbf{Y}}_L)$, respectively. The targets in graph learning with weak structure, weak features, and weak labels scenarios are to predict the labels of unlabeled nodes $\mathcal{V}_{U}$ with $\mathcal{D}_{ws}$, $\mathcal{D}_{wf}$, and $\mathcal{D}_{wl}$ for model training, respectively. These three scenarios are defined as basic GLWI scenarios. 
\end{definition}

In the real world, the data deficiencies often occur, more or less, in three aspects simultaneously, leading to the more intractable extreme GLWI scenario:

\begin{definition}[Extreme GLWI scenario] \label{def:x}
Let graph data with extremely weak information be $\mathcal{D}_{x}=(({\mathcal{V}}, \check{\mathcal{E}}, \check{\mathbf{X}}), \check{\mathbf{Y}}_L)$. The target in extreme GLWI scenario is to predict the labels of unlabeled nodes $\mathcal{V}_{U}$ with $\mathcal{D}_{x}$ for model training. 
\end{definition}

Notably, in basic scenarios, the graph data only has one type of weak information; on the contrary, the structure, features, and labels are all deficient in extreme scenario. Due to the mutual effects among different data deficiency, extreme scenario is \textbf{more challenging} than basic scenarios.

%% file: 4_moti.tex
\begin{figure*}[!t]
 \centering
 \subfigure[Graph learning with weak features]{
   \includegraphics[height=0.18\textwidth]{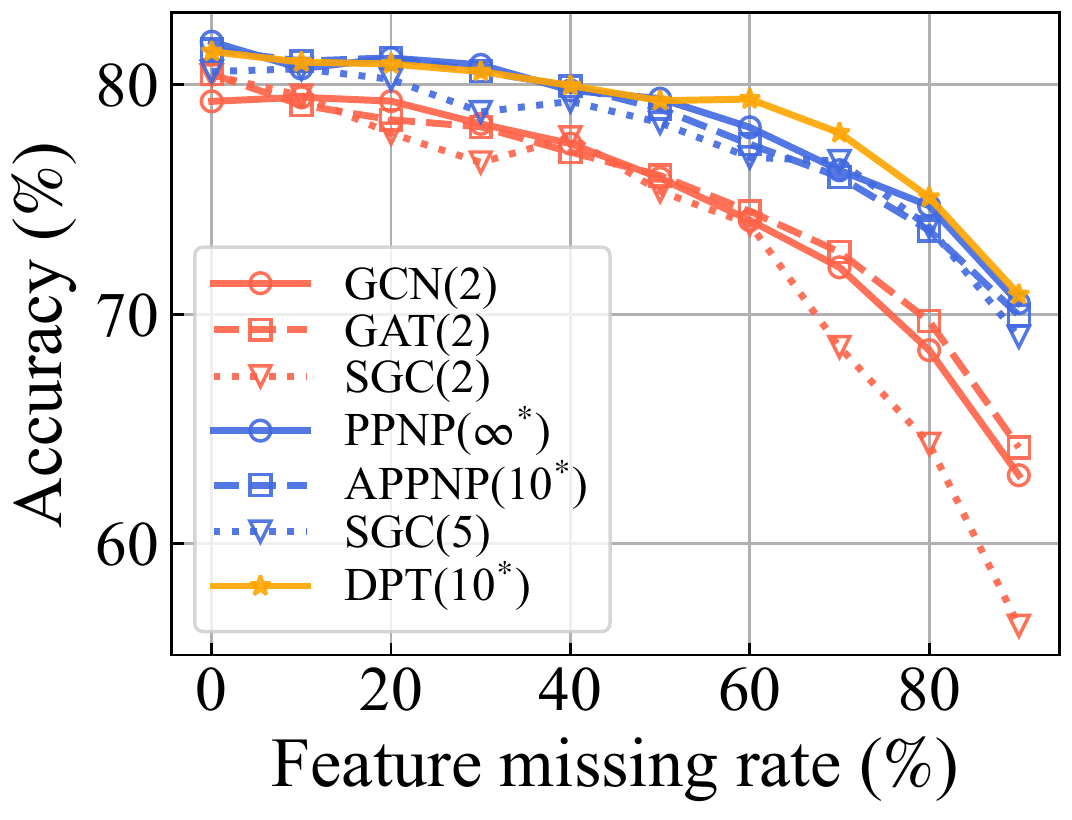}
   \label{subfig:wf_propstep}
 }
 \subfigure[Graph learning with weak labels]{
   \includegraphics[height=0.18\textwidth]{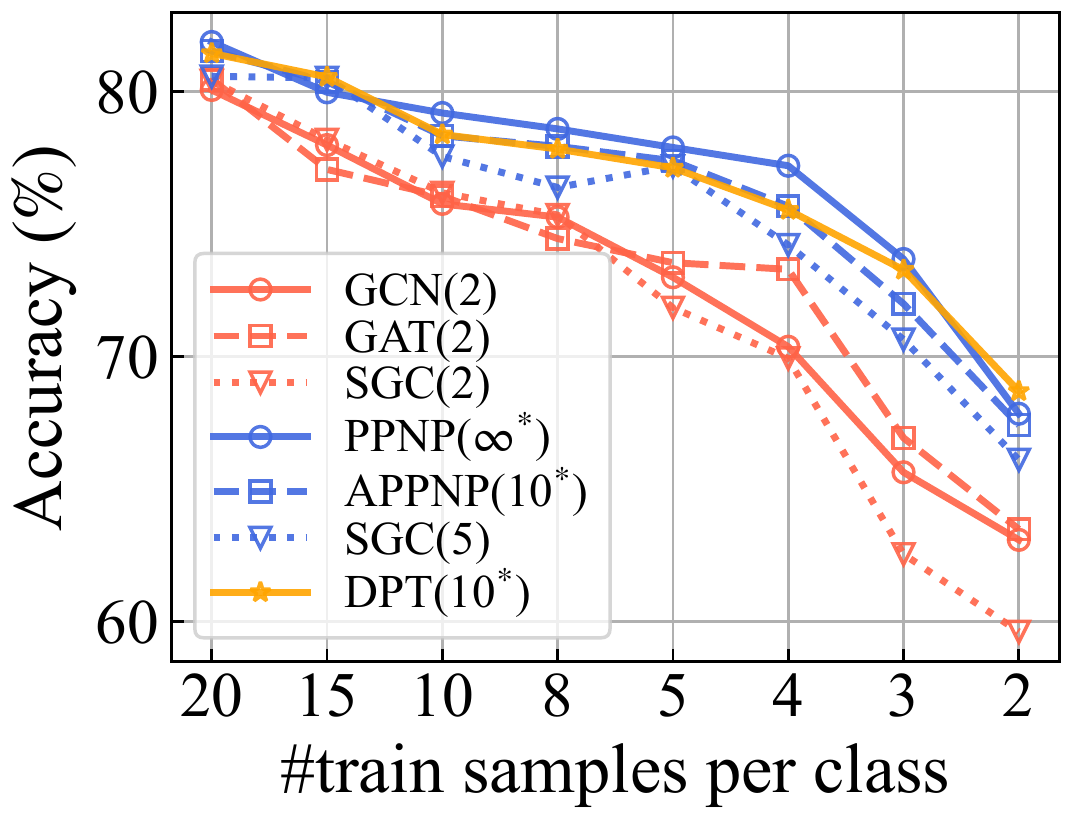}
   \label{subfig:wl_propstep}
 }
  \subfigure[Graph learning with weak structure]{
   \includegraphics[height=0.18\textwidth]{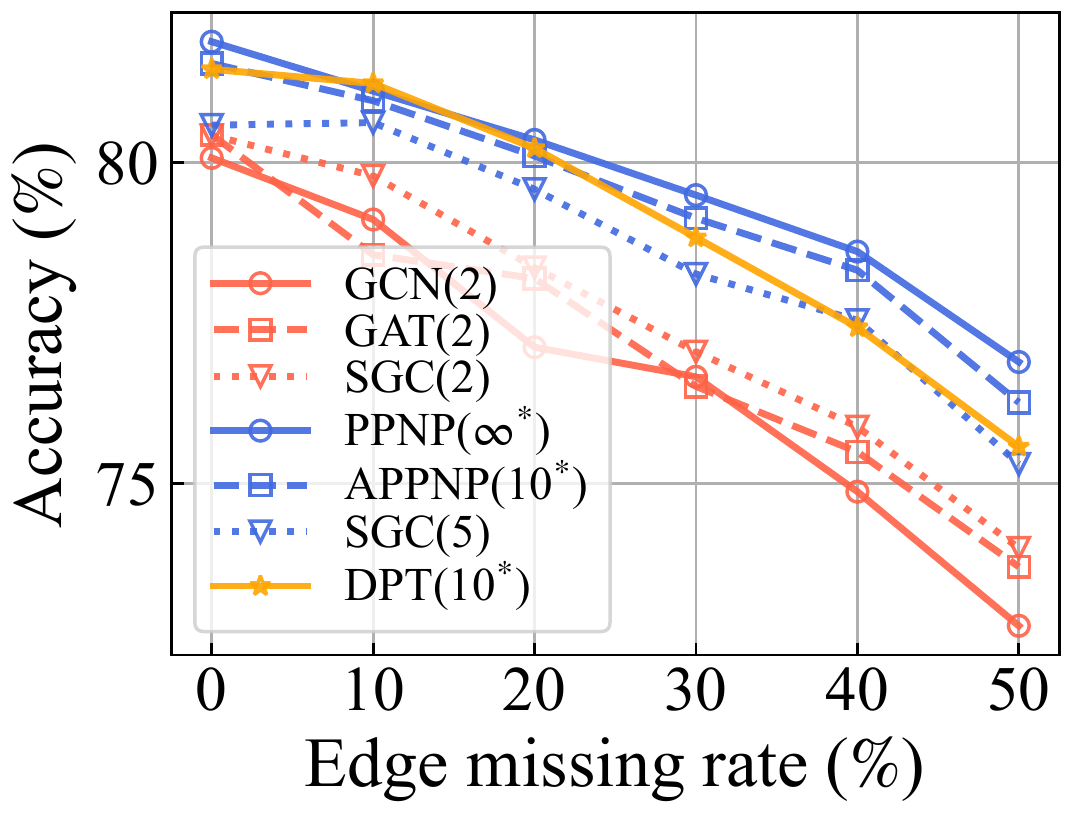}
   \label{subfig:ws_propstep}
 } 
 \subfigure[Combinations of GLWI scenarios]{
   \includegraphics[height=0.18\textwidth]{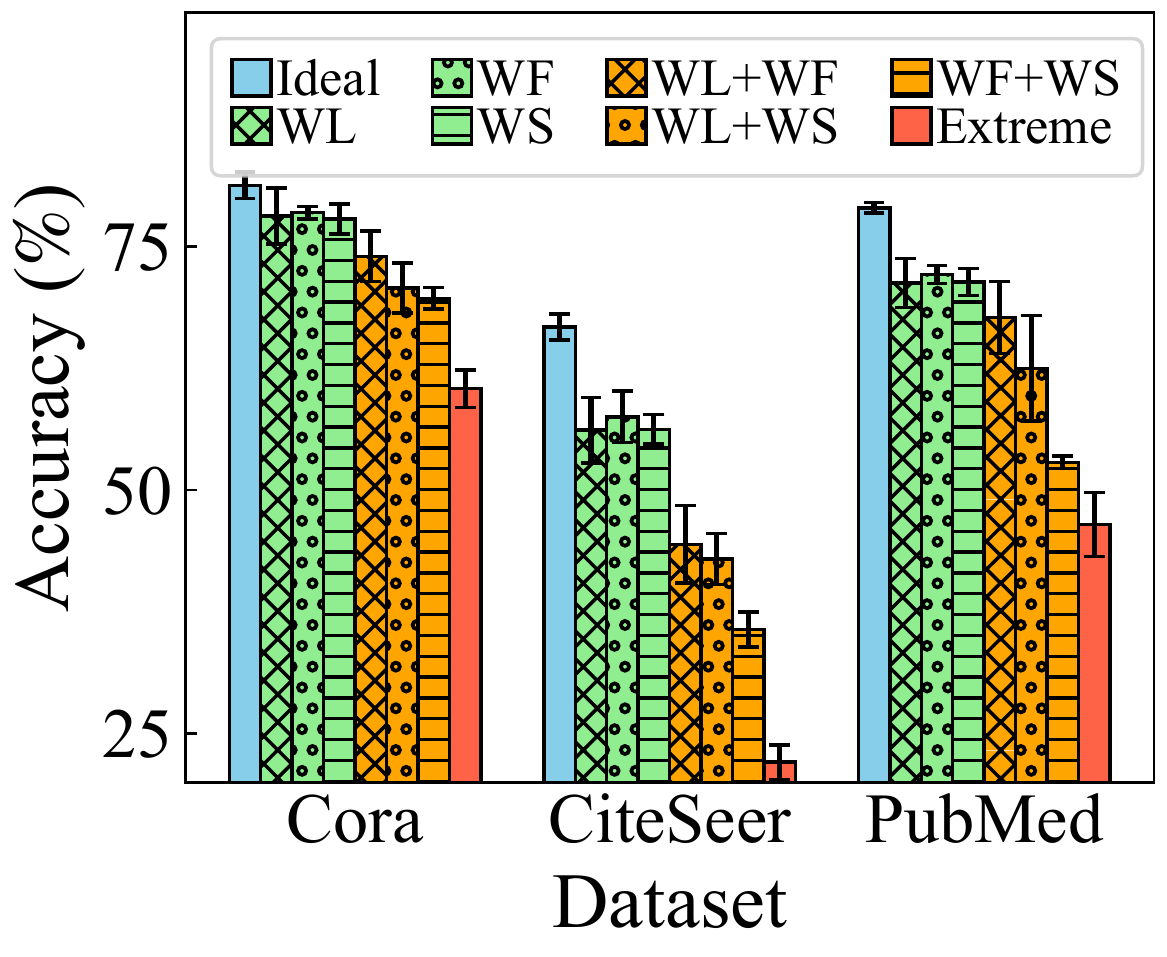}
   \label{subfig:combine_scenario}
 }
 \vspace{-0.4cm}
 \caption{The performance comparison of (a-c) different GNNs in basic GLWI scenarios on Cora dataset, where numbers in the brackets indicate $s_p$ and $*$ indicates graph diffusion-based model, and (d) different combinations of GLWI scenarios.}
 \vspace{-0.3cm}
 \label{fig:combine_scenario}
\end{figure*}

In this section, we expose the key to solving the GLWI problem is to execute \textit{effective information propagation} in GNNs. Firstly, we discuss the critical roles of information propagation in handling graph data with weak information. Then, with empirical analysis, we find two crucial criteria that enable effective information propagation and hence benefit GLWI, i.e., employing long-range propagation and alleviating the stray node problem.  

\subsection{Roles of Information Propagation in GLWI} \label{subsec:role}

In GNNs, propagation is a fundamental operation that transmits information along edges in graph-structured data~\cite{ding2022data,ding2023eliciting}. 
It is noteworthy that, from the perspectives of features, labels, and structure, information propagation respectively plays unique yet pivotal roles in learning with deficient data. 
In the following paragraphs, we will discuss how information propagation leverages the features, labels, and structure in graph learning, especially when graph data is incomplete and insufficient. 

\noindent\textbf{Role in handling weak features.} 
From the perspective of features, the information propagation in GNNs can naturally \textit{complete the features} with contextual knowledge~\cite{ding2022data,liu2021tail}. 
For example, in a citation network, a target node (paper) has features (keywords) ``BERT'' and  ``Text''. Through propagation, the features from neighbors (e.g., ``GPT'', ``XLNet'', and ``Sentence'') can provide supplementary information to complete the features of the target node, which makes the model easier to classify this node as an ``NLP paper''. 
For graph data with weak features, such a ``propagate to complete'' mechanism becomes more significant, since the missing features are in urgent need of complement from contextual features. 

\noindent\textbf{Role in handling weak labels.} 
In semi-supervised node classification tasks, information propagation also plays a unique role, i.e., \textit{spreading the supervision signals} from the labeled nodes to the unlabeled nodes~\cite{ding2022meta}. According to the influence theorem in~\cite{wang2021combining}, in GCN~\cite{kipf2017semi}, the label influence of a labeled node $v_l$ on an unlabeled node $v_u$ equals to the expectation of the cumulative normalized feature influence of $v_l$ on $v_u$ in the scope of reception field. That is to say, along with information propagation, a labeled node can influence all the unlabeled nodes within $s_p$ hops, spreading the supervision signals to these nodes. For graph data where labeled nodes are extremely scarce, we need effective information propagation to allow more unlabeled nodes to be covered by the correct supervision signals from labeled nodes. 

\noindent\textbf{Role in handling weak structure.} 
During information propagation, the edges in graph structures are the ``bridges'' to \textit{communicate knowledge} between adjacent nodes. 
However, when edges are partly missing, communication becomes more difficult due to the lack of bridges. 
In this case, an ideal information propagation strategy should enable efficient knowledge communication by fully leveraging the existing edges. 
For instance, if a key connection that links two nodes is broken, effective information propagation can leverage their long-range dependency to preserve the communication between these nodes, relieving the impact of missing edges. 

Notably, graph structure provides the channel for the propagation of contextual features and supervision signals, meaning that its quality can also affect feature imputation and label transmission. 
Hence, improving the quality of incomplete structure is significant in GLWI, especially with incomplete features and labels. 

\vspace{-0.15cm}
\begin{tcolorbox}
[boxsep=0mm,left=2.5mm,right=2.5mm,colframe=black!55,colback=black!5]
\textbf{Summary:} In this subsection, we illustrate that effective information propagation helps feature completion, supervision signal spreading, and knowledge communication. In this case, a follow-up \textit{\textbf{Motivated Question}} is: \textit{``How to make information propagation more effective {under the setting of GLWI}?}''
\end{tcolorbox}
\vspace{-0.15cm}

\subsection{Long-Range Propagation Benefits GLWI} \label{subsec:large_sp} 
Without modifying the network architectures and propagation mechanisms, an effective solution to the \textit{\textbf{Motivated Question}} is utilizing \textbf{long-range propagation}, i.e., enlarging the propagation step $s_p$ in GNNs. 
From the perspectives of features, labels, and structure, long-range propagation can generally relieve data deficiency. In graph learning with weak features, long-range propagation allows the GNN models to consider a wider range of contextual nodes, which provides ampler contextual knowledge for feature imputation. On graph data where labeled nodes are scarce, GNNs with larger $s_p$ can broadcast the rare supervision signals to more unlabeled nodes. 
When graph structure is incomplete, a larger $s_p$ enables long-range communication between two distant nodes, which better leverages the existing edges.

\noindent\textbf{Empirical analysis.} 
To expose the impact of long-range propagation in GLWI, we run 3 \textcolor[RGB]{238,112,97}{small-$s_p$ models} and 3 \textcolor[RGB]{90,122,221}{large-$s_p$ models} on three GLWI scenarios with varying degrees of data deficiency. More experimental details are provided in Appendix~\ref{appendix:sp_exp}. 
From Fig.~\ref{subfig:wf_propstep}-\ref{subfig:ws_propstep}, we have the following observations: 1) On three scenarios, all the large-$s_p$ models generally outperform the small-$s_p$ models by significant margins, which verifies the effectiveness of large-$s_p$ models in GLWI scenarios; 2) with data incompleteness getting more severe, the performance gaps between large-$s_p$ and small-$s_p$ models become more remarkable, which demonstrates the superiority of large-$s_p$ models in learning from severely deficient data; 3) keeping other hyper-parameters unchanged, SGC($s_p=5$) achieves significantly better performance than SGC($s_p=2$), indicating that the larger $s_p$ is the major cause for the performance gain; 4) among large-$s_p$ models, the GNNs with graph diffusion mechanism tend to perform better. 
To sum up, our empirical analysis verifies that GNNs with larger $s_p$ can generally perform better on basic GLWI scenarios;  

\noindent\textbf{Discussion.}
1) Based on homophily assumption~\cite{zhu2020beyond, liu2022beyond, zheng2022graph, zheng2023finding}, we deduce that nodes with similar features/labels tend to be connected in graph topology, supporting the effectiveness of long-range propagation in feature imputation and supervision signal spreading. However, if $s_p$ is overlarge, noisy information would inevitably occur in receptive fields, which may degrade the performance. During grid search, we also find that SGC($s_p=10$) performs worse than SGC($s_p=5$), indicating that $s_p$ should be kept within an appropriate range. Here, we would like to point out that the default $s_p=2$ for most GNNs is usually ineffective for GLWI. 2) A large $s_p$ tends to aggravate the over-smoothing issue~\cite{li2018deeper, zhang2022graph}. In our experiments, we find that graph diffusion mechanism can alleviate the issue by adding ego information at each propagation step with residual connection~\cite{gasteiger2018predict, gasteiger2019diffusion}, which allows larger $s_p$ while preserving high performance. 

\vspace{-0.15cm}
\begin{tcolorbox}[boxsep=0mm,left=2.5mm,right=2.5mm,colframe=black!55,colback=black!5]
\textbf{Summary:} With empirical analysis, we {derive} \textit{\textbf{Criterion 1} {to handle GLWI problem}: enabling long-range propagation leads to effective information propagation, which further alleviates the deficiency in structure, features, and labels.}     
\end{tcolorbox}
\vspace{-0.15cm}

\subsection{Stray Nodes Hinder GLWI} 
\label{subsec:cc_graph}

\begin{figure}[!t]
 \centering
 \vspace{-0.2cm}
 \subfigure[Weak structure leads to stray nodes]{
   \includegraphics[height=0.46\columnwidth]{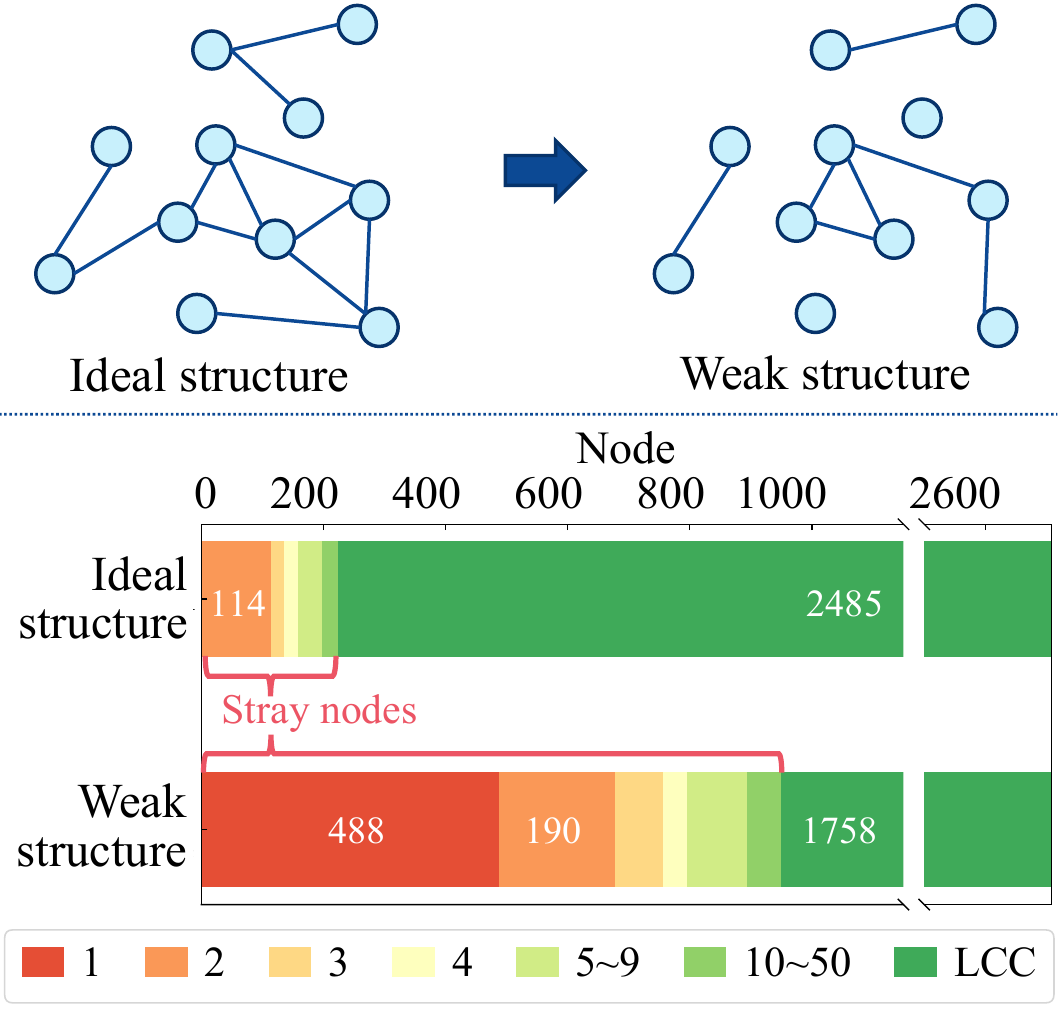}
   \label{subfig:cc_distribution}
 }  \hspace{1mm}
 \subfigure[Stray nodes hinder feature completion and supervision signal spreading]{
   \includegraphics[height=0.46\columnwidth]{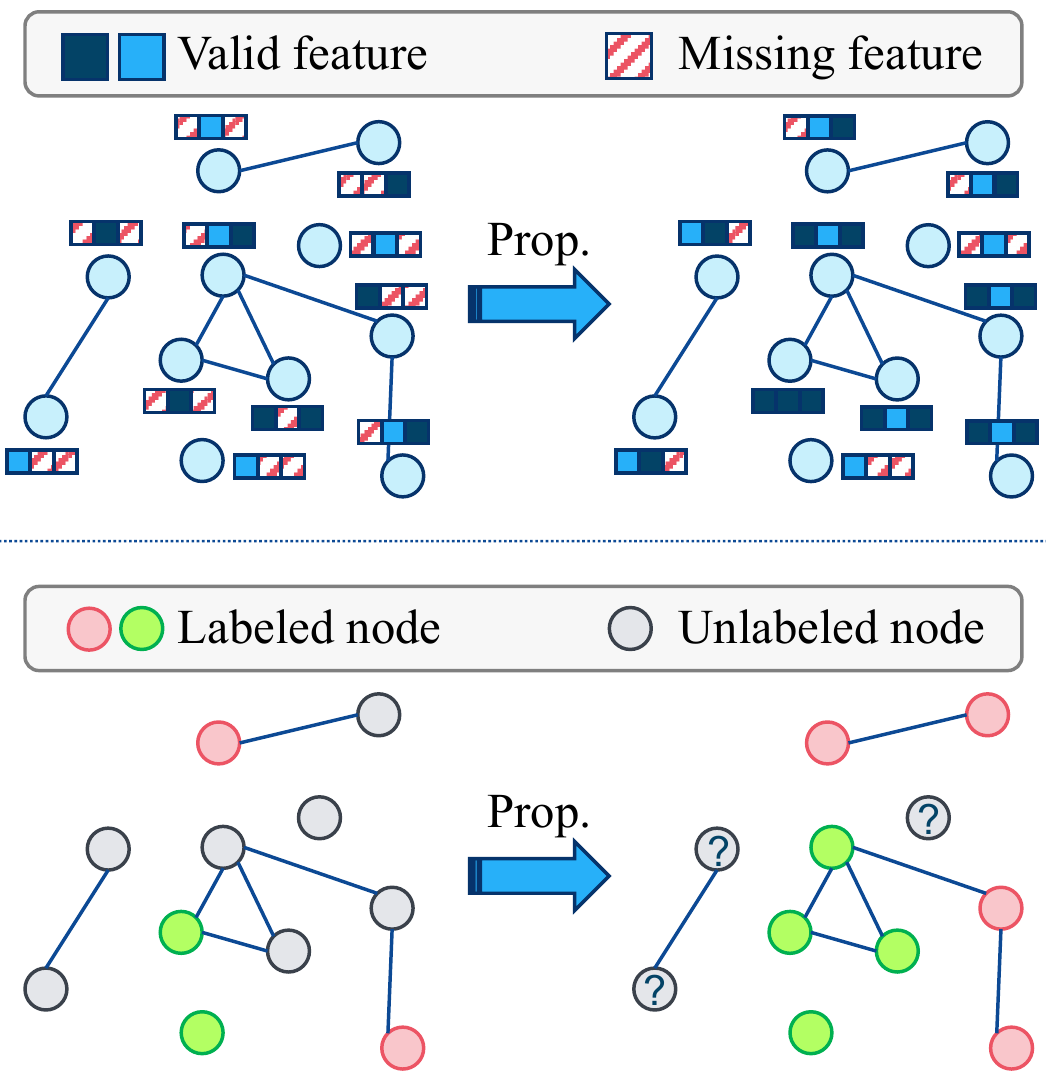}
   \label{subfig:prop_limit}
 }
 \vspace{-0.4cm}
 \caption{Sketch maps to illustrate stray node problem.}
 \vspace{-0.4cm}
 \label{fig:method_moti}
\end{figure}

In the above subsection, we demonstrate that larger $s_p$ leads to effective information propagation and hence benefits basic GLWI scenarios. 
Then, we are curious about \textit{how do large-$s_p$ models perform in the scenarios where data insufficiency in features/labels/structure are entangled?} 
To answer this question, we conduct experiments (detailed settings are in Appendix~\ref{appendix:comb_exp}) to compare the performance of APPNP($s_p=20$) on graph data with different combinations of weak features (WF), weak labels (WL), and weak structure (WS). 
From the results in Fig.~\ref{subfig:combine_scenario}, we witness sharp decreases in performance when data are with multiple types of deficiencies. Even with a larger propagation step, unfortunately, conventional GNNs still suffer from the extremely weak information in graph data.

Based on the results, one might ask: {does larger $s_p$ make information propagation effective enough for extreme GLWI scenario?} 
Recalling that graph structure provides the ``bridges'' for information propagation, we speculate the quality of graph structure can also affect the effectiveness of information propagation. Some clues can be found in Fig.~\ref{subfig:combine_scenario}: among the two-aspect combination scenarios, ``WF+WS'' and ``WL+WS'' have severer performance degradation than ``WF+WL'', which indicates the incomplete graph structure (WS) is the major factor hindering GLWI.

To understand how weak structure hinders GLWI, we first investigate the difference between ideal structures and weak structures\footnote{In the case study, we construct weak structures by randomly removing $50\%$ of edges.}. By comparing the distributions of node connection in ideal and incomplete structures, we have an interesting observation: as shown in the upper of Fig.~\ref{subfig:cc_distribution}, in ideal structures, the vast majority of nodes are connected together to form the largest connected component (LCC); in contrast, in weak structures, there exist more stray subgraphs composed of few nodes and more isolated nodes that are even independent to other nodes. On Cora dataset, we visualize the distribution of nodes from connected components with different sizes in the lower of Fig.~\ref{subfig:cc_distribution}. We can demonstrate that in ideal structure, over $90\%$ ($2485$ out of $2708$) of nodes are included in LCC, while this percentage decreases to $64.9\%$ in weak structure. Moreover, about $18\%$ of nodes in weak structure are isolated, while no isolated node exists in ideal structure. 

For simplicity, we denote \textit{stray nodes} as the nodes from stray subgraphs and the isolated nodes, and denote \textit{LCC nodes} as the nodes within LCC. With empirical analysis, we find that in extreme GLWI scenario, information propagation is ineffective on stray nodes, leading to sub-optimal performance in extreme GLWI scenario.

\input{tables/lcc_vs_stray.tex}

\noindent\textbf{Stray nodes hinder feature completion.} 
In Sec.~\ref{subsec:role}, we illustrate that GNNs are able to complete the features with contextual knowledge via recurrent propagation. However, for stray nodes, the missing information is hard to be filled with limited contextual nodes. 
As shown in the upper of Fig.~\ref{subfig:prop_limit}, if a specific feature is missing in all nodes within a stray subgraph, it cannot be completed by propagation, even if we increase $s_p$. 
For the isolated nodes, the propagation is unable to complete their features, which is a worse case. In Table~\ref{tab:lcc_vs_stray}, we show the average L2 distance between raw features and the features after propagation in SGC~\cite{wu2019simplifying}. We find that the distances on LCC nodes are 2x\textasciitilde9x larger than those on stray nodes, demonstrating that LCC nodes receive much more completion than stray ones.

\noindent\textbf{Stray nodes hinder supervision signal spreading.} 
In Sec.~\ref{subsec:role}, we point out that GNNs also play the role of spreading supervision signals from labeled nodes to unlabeled nodes. Unfortunately, if all the nodes in a stray subgraph are unlabeled, the supervision single is hard to reach them via propagation (e.g., the nodes with ``?'' in the lower of Fig.~\ref{subfig:prop_limit}). Meanwhile, for the labeled stray nodes, the supervision signals are also trapped in the small connected components and then cannot be propagated to most nodes. In this case, when labeled nodes are extremely scarce, a large number of nodes would be out of the coverage of supervision. As illustrated in Table~\ref{tab:lcc_vs_stray}, on APPNP~\cite{gasteiger2018predict}, the test accuracy on LCC nodes is $13.4\%$ \textasciitilde $60.5\%$ higher than the stray nodes, which indicates that stray nodes are more potential to be misclassified due to the lack of supervision. 

\noindent\textbf{Challenge.} 
Although the stray nodes are easy to be identified in an incomplete graph, it is of great difficulty to handle stray nodes in GLWI. 
A feasible solution is to connect the stray nodes to LCC. However, directly linking irrelevant nodes together may introduce noisy edges to the original graph, which further harms the feature imputation and label spreading. Moreover, when features and labels are deficient, it is hard to determine how to build the connections between the stray nodes and other nodes. 

\vspace{-2mm}
\begin{tcolorbox}[boxsep=0mm,left=2.5mm,right=2.5mm,colframe=black!55,colback=black!5]
\textbf{Summary:} By investigating how incomplete structure affect feature imputation and supervision signal spreading, we can summarize \textit{\textbf{Criterion 2}: the key to handling extreme GLWI scenario is to address the ineffective information propagation problem on the stray nodes isolated from LCC}.  
\end{tcolorbox}

%% file: tables/lcc_vs_stray.tex
\begin{table}[t]
\centering
\vspace{-0.1cm}
\caption{Comparison between LCC and stray nodes w.r.t. feature-wise distance/accuracy in extreme GLWI scenario. Experimental details please find in Appendix~\ref{appendix:stray_exp}.}
\label{tab:lcc_vs_stray}
\vspace{-0.2cm}
\resizebox{0.95\columnwidth}{!}{
\begin{tabular}{p{1.3cm}|p{1.8cm}<{\centering} p{1.8cm}<{\centering}|p{1.8cm}<{\centering} p{1.8cm}<{\centering}}
\toprule
\multirow{2}*{{{Dataset}}} & \multicolumn{2}{c} {L2 distance} & \multicolumn{2}{c} {Test accuracy}  \\
\cline{2-3} \cline{4-5}
 & LCC nodes & stray nodes & LCC nodes & stray nodes  \\
\hline
Cora & $2.7870$ & $1.0889$ & $62.79$ & $39.12$ \\
CiteSeer & $3.7238$ & $1.7312$ & $56.82$ & $35.99$  \\
PubMed & $0.2553$ & $0.0278$ & $67.54$ & $59.93$ \\
\bottomrule
\end{tabular}}
\vspace{-0.5cm}
\end{table}

%% file: 5_method.tex
From the analysis in Sec.~\ref{sec:base}, we pinpoint that the key to addressing GLWI problem is to enable effective information propagation. To this end, we can design GNN models for GLWI following two crucial criteria: 
\textit{\textbf{Criterion 1} -  enabling long-range propagation} and \textit{\textbf{Criterion 2} -  handling stray node problem}. With the guidance of \textit{\textbf{Criterion 1}}, in this section, we first present a strong base model termed \ourbase, a large-$s_p$ GNN that balances the effectiveness and efficiency. Then, following \textit{\textbf{Criterion 2}}, we further propose \ourmethod by introducing a dual-channel architecture with an augmented global graph, which relieves the stray node problem. 

\subsection{Diffused Propagation then Transformation} \label{subsec:base}

Motivated by \textit{\textbf{Criterion 1}}, enlarging the propagation step $s_p$ is critical for effective information propagation; however, the growing computational complexity with the increase of $s_p$ is also non-negligible. 
For entangled GNNs (e.g., GCN~\cite{kipf2017semi} and GAT~\cite{velivckovic2018graph}) where transformation and propagation operations are coupled, enlarging $s_p$ leads to the growing complexities of both types of operations. Moreover, stacking numerous entangled GNN layers may also result in model degradation, affecting the model performance~\cite{zhang2022model}. 
Even for some disentangled GNNs (i.e., \texttt{TTPP} models such as PPNP/APPNP~\cite{gasteiger2018predict}), the complexity of propagation operation still grows with $s_p$, which also causes heavy computing cost when $s_p$ is large. 
Fortunately, some \texttt{PPTT} models~\cite{wu2019simplifying} can be efficient in propagating since they only propagate once during the training procedure. 
Inspired by the efficiency of \texttt{PPTT} models and the effectiveness of graph diffusion mechanism (discussed in Sec~\ref{subsec:large_sp}), we present \textit{\textbf{D}iffused \textbf{P}ropagation then \textbf{T}ransformation} (\ourbase for short), a strong base GNN model for GLWI, that serves as the backbone of our proposed method.

On the highest level, \ourbase is composed of two steps: graph diffusion-based propagation and MLP-based transformation. Concretely, graph diffusion-based propagation is built upon the approximated topic-sensitive PageRank diffusion~\cite{gasteiger2018predict} and is written as: \looseness-2
\begin{equation}
\label{eq:prop}
\begin{aligned}
\mathbf{X}^{(0)}=&\mathbf{X}, \\
\mathbf{X}^{(\mathrm{t}+1)}=&(1-\alpha) \tilde{\mathbf{A}} \mathbf{X}^{(\mathrm{t})}+\alpha \mathbf{X}, \\
\overline{\mathbf{X}}=&\mathbf{X}^{(T)},
\end{aligned}
\end{equation}
\noindent where $\alpha \in (0,1]$ is the restart probability, $\tilde{\mathbf{A}}$ is the normalized adjacency matrix, $\overline{\mathbf{X}}$ is the propagated feature matrix, $T$ is the iteration number of propagation (equal to $s_p$ in \ourbase), and $t \in [0, T-1]$. 

After graph diffusion-based propagation, MLP-based transformation maps $\overline{\mathbf{X}}$ into the predicted label matrix $\overline{\mathbf{Y}}$:
\begin{equation}
\label{eq:trans}
\begin{aligned}
\mathbf{H}=&\texttt{MLP}_{1}(\overline{\mathbf{X}}) = \sigma(\overline{\mathbf{X}}\mathbf{W}_{1}), \\
\overline{\mathbf{Y}}=&\texttt{MLP}_2({\mathbf{H}}) = \operatorname{softmax}(\mathbf{H}\mathbf{W}_2),
\end{aligned}
\end{equation}
\noindent where $\mathbf{H} \in \mathbb{R}^{n \times e}$ is the representation matrix, and $\mathbf{W}_{1}$ and $\mathbf{W}_{2}$ are learnable parameter matrices. After that, a cross-entropy loss $\mathcal{L}_{ce}$ for model training can be computed from $\overline{\mathbf{Y}}$.

In contrast to conventional GNNs, \ourbase has several nice properties, including competitive effectiveness in basic GLWI scenarios (see Fig.~\ref{subfig:wf_propstep}-\ref{subfig:ws_propstep}), 
high running efficiency, and flexible memory requirement. More discussion and comparison about \ourbase are conducted in Appendix~\ref{appendix:dpt_discuss}. 

\subsection{Dual-Channel DPT}

\begin{figure}[!t]
\centering
  \includegraphics[width=1\columnwidth]{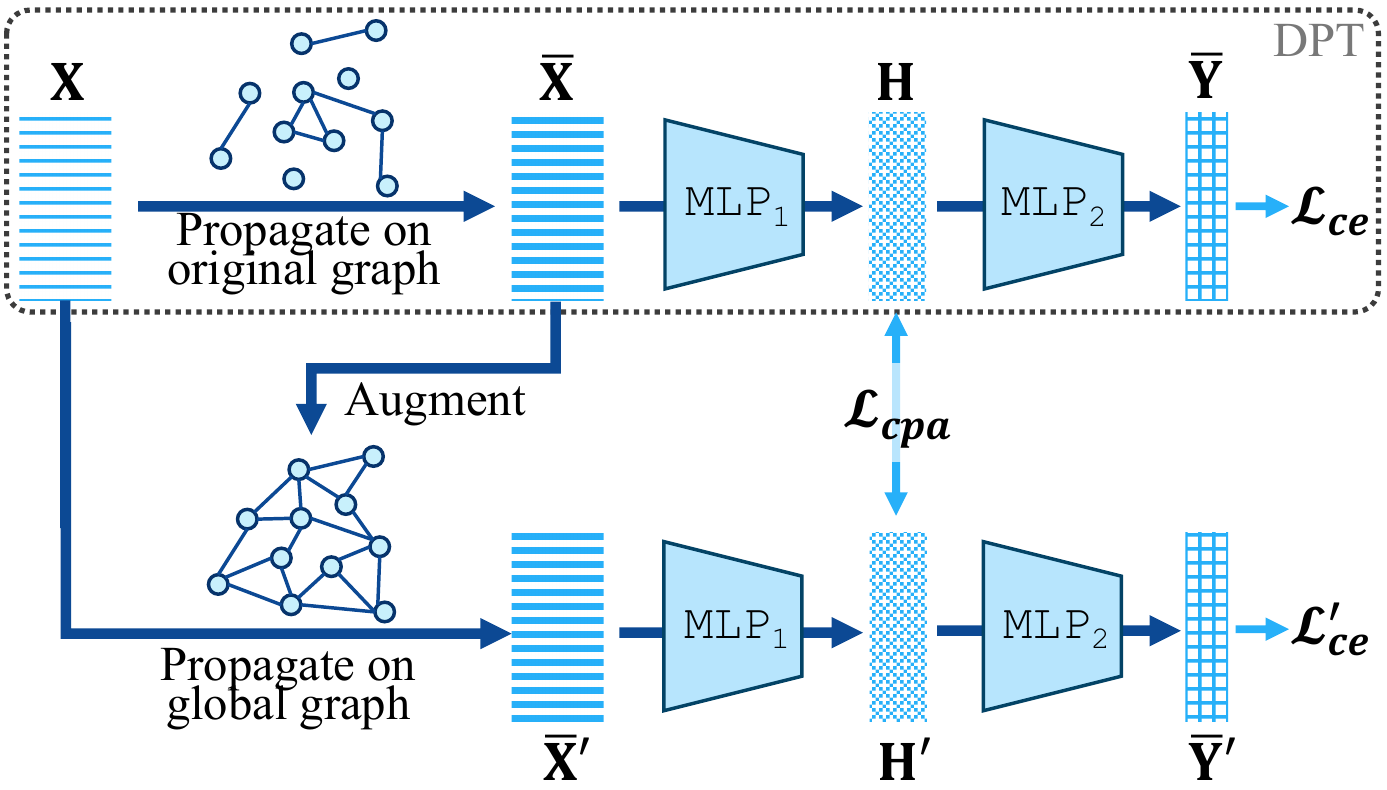}
  \vspace{-0.4cm}
  \caption{Overall framework of \ourmethod.
  }
  \label{fig:pipeline}
  \vspace{-0.3cm}
\end{figure}

With the guidance of \textit{\textbf{Criterion 2}}, the key to learning on graph data with extremely weak information is to address the stray node problem. 
To this end, based on \ourbase, we develop a novel method termed \textit{\textbf{D}ual-channel \textbf{D}iffused \textbf{P}ropagation then \textbf{T}ransformation} (\ourmethod for short). The core motivation behind \ourmethod is to construct a global graph where all nodes are connected, and then jointly train \ourbase on both the original and global graphs. Through the information propagation on global graph, stray nodes can also receive sufficient knowledge, which eliminates the stray node problem. 
As shown in Fig.~\ref{fig:pipeline}, from \ourbase to \ourmethod, the main improvements are three-fold: 1) generating an augmented global graph from propagated feature $\overline{\mathbf{X}}$; 2) running \ourbase on the global graph in another channel; and 3) regularizing the representations of two channels with a contrastive prototype alignment loss $\mathcal{L}_{cpa}$. The following paragraphs illustrate these crucial designs respectively.

\noindent\textbf{Dual-Channel Training with Global Graph.} 
Recalling the degradation caused by stray node problem, the bottleneck is that the stray nodes have limited adjacent nodes to connect with. To solve this problem, a natural and straightforward solution is to introduce a new graph where all nodes have sufficient adjacent nodes. However, directly replacing the original graph with a new one may lead to inevitable information loss. Alternatively, we propose to generate an augmented global graph alongside the original graph, and then extract knowledge from the global graph during model training. 

In \ourmethod, we employ k-nearest neighbor (kNN) graph as the global graph, which ensures that each node has at least $k$ neighbors. To leverage the features completed by \ourbase, we construct kNN graph from the propagated features matrix $\overline{\mathbf{X}}$ instead of raw features $\mathbf{X}$. In concrete, the kNN adjacency matrix is written by:

\begin{equation}
\label{eq:knn}
\mathbf{A}' \text{, where } \mathbf{A}'_{ij}= \left\{\begin{aligned}
&1,  \!   & \operatorname{s}(\overline{\mathbf{X}}_{i}, \overline{\mathbf{X}}_{j}) \geq \operatorname{min}(\tau(\overline{\mathbf{X}}_{i}, k), \tau(\overline{\mathbf{X}}_{j}, k)), \\
&0,   \!   & \text{otherwise},
\end{aligned}\right.
\vspace{-1mm}
\end{equation}

\noindent where $\operatorname{s}(\overline{\mathbf{X}}_{i}, \overline{\mathbf{X}}_{j})$ is the similarity between vectors $\overline{\mathbf{X}}_{i}$ and $\overline{\mathbf{X}}_{j}$, and $\tau(\overline{\mathbf{X}}_{i},k)$ returns the similarity between $\overline{\mathbf{X}}_{i}$ and its $k$-th similar row vector in $\overline{\mathbf{X}}$. Here $\mathbf{A}'$ is inherently symmetric.  

Then, in a parallel manner, we execute \ourbase on the original graph as well as the global graph simultaneously. For the original channel, the computation follows Sec.~\ref{subsec:base}. For the global channel, by replacing the adjacency matrix in Eq.~(\ref{eq:prop}) to  $\tilde{\mathbf{A}}'=\mathbf{D}'^{-1/2}\mathbf{A}'\mathbf{D}'^{-1/2}$, we calculate the global propagated features $\overline{\mathbf{X}}'$. After acquiring the output $\overline{\mathbf{Y}}'$ via Eq.~(\ref{eq:trans}), we can finally compute the loss function $\mathcal{L}'_{ce}$ at the global channel. By training the shared-weight MLP model with $\mathcal{L}_{ce}$ and $\mathcal{L}'_{ce}$ jointly, \ourmethod is able to capture informative knowledge from two graph views and alleviate the affect by stray nodes. Since $\overline{\mathbf{X}}'$ can also be pre-computed before training, \ourmethod inherits the high efficiency of \ourbase.

\noindent\textbf{Contrastive Prototype Alignment.} 
Now we can train the backbone \ourbase model on both the original and global views. However, the naive dual-channel pipeline has two limitations. First, although original and global channels share model parameters, the generated representations, especially those of stray nodes, can be significantly different due to input structure differences. Consequently, the model may fail to capture common knowledge from both views or even be confused by the disordered supervision signals. Second, both $\mathcal{L}_{ce}$ and $\mathcal{L}'_{ce}$ are computed based on labeled nodes, leading to the potential over-fitting problem when training samples are scarce. 

To bridge the gaps, we introduce a contrastive prototype alignment loss that enhances the semantic consistency between two channels and, at the same time, extracts supervision signals from unlabeled samples. At the first step, we employ a linear projection layer to map the representations $\mathbf{H}$ and $\mathbf{H}'$ into latent embeddings $\mathbf{Z}=\mathbf{H}\mathbf{W}_{3}$ and $\mathbf{Z}'=\mathbf{H}'\mathbf{W}_{3}$. Then, for each class $j \in [1, \cdots, c]$, we acquire its prototype~\cite{tanfederated, tan2022fedproto} by calculating the weighted average of latent embeddings: 

\begin{equation}
\label{eq:proto}
\mathbf{p}_j=\sum\nolimits_{\underset{i}{\operatorname{argmax}}(\overline{\mathbf{Y}}_i)=j} \frac{s_i Z_i}{S_j} , \quad \mathbf{p}'_j=\sum\nolimits_{\underset{i}{\operatorname{argmax}}(\overline{\mathbf{Y}}_i)=j} \frac{s_i Z'_i}{S_j},
\end{equation}

\noindent where weight $s_i=1$ for labeled nodes and $s_i=\max ({\overline{Y}}_i)$ (i.e., the confidence in prediction) for unlabeled nodes and $S_j$ is the sum of weight $s_i$ of all nodes allocated to class $j$. Once the prototypes are computed, we regularize the prototypes from two channels with an Info-NCE-based~\cite{chen2020simple} contrastive prototype alignment loss:

\begin{equation}
\label{eq:cpa}
\mathcal{L}_{cpa}=-\frac{1}{2 c} \sum_{j=1}^c\left(\log \frac{\mathrm{f}(\mathbf{p}_j, \mathbf{p}_j')}{\sum_{q \neq j} \mathrm{f}(\mathbf{p}_j, \mathbf{p}_q')}+\log \frac{\mathrm{f}(\mathbf{p}_j, \mathbf{p}_j')}{\sum_{q \neq j} \mathrm{f}(\mathbf{p}_q, \mathbf{p}_j')}\right),
\end{equation}

\noindent where $\mathrm{f}(\mathbf{a},\mathbf{b})=\mathrm{e}^{{\cos \left(\mathbf{a},\mathbf{b}\right)}/{\tau}}$, $\cos(\cdot,\cdot)$ is the cosine similarity function, and $\tau$ is the temperature hyper-parameter. {$\mathcal{L}_{cpa}$ increases the agreement between the representations of original and global views on labeled and unlabeled nodes, which helps the model distill informative knowledge from the global view, especially for the stray nodes. Moreover, compared to traditional sample-wise contrastive loss~\cite{chen2020iterative}, $\mathcal{L}_{cpa}$ enables the model to learn class-level information and further leverage labels, and also enjoys much lower time complexity ($\mathcal{O}(ec^2)$ rather than $\mathcal{O}(en^2)$)}. 

\noindent\textbf{Learning Objective.} 
By combining the above three losses with trade-off coefficients $\gamma_1$ and $\gamma_2$, the overall objective of \ourmethod is:

\begin{equation}
\label{eq:loss}
\mathcal{L} = \mathcal{L}_{ce} + \gamma_1\mathcal{L}'_{ce} + \gamma_2\mathcal{L}_{cpa}.
\end{equation}

\noindent\textbf{Scalability Extension.} 
In order to adapt \ourmethod to large-scale graphs efficiently, we introduce the following mechanisms. 

1) To reduce the cost of computing kNN graphs, based on locality-sensitive approximation~\cite{fatemi2021slaps,halcrow2020grale}, we design a glocal approximation algorithm to efficiently construct globally connected kNN graphs. The core idea is to approximate local kNN twice with different batch splits and integrate two kNN graphs into a connected global graph. Detailed algorithm please see Appendix~\ref{appendix:fast_knn}. 

2) To reduce the computational cost of training phase, we adopt a mini-batch semi-supervised learning strategy. 
In each epoch, we only sample a batch of unlabeled nodes $\mathcal{V}_B$ ($|\mathcal{V}_B|=n_B \ll n$) along with labeled nodes $\mathcal{V}_L$ for model training. In this case, $\mathcal{L}_{cpa}$ is calculated on $\mathcal{V}_L \cup \mathcal{V}_B$ instead of all nodes, which significantly saves the requirement on memory and time. 
Finally, the time complexity per training epoch is $\mathcal{O}(e(n_L+n_B)(e+d+c)+c^2e)$. The detailed complexity analysis is given in Appendix~\ref{appendix:complexity}, and the overall algorithm of \ourbase is in Appendix~\ref{appendix:algo}.

%% file: 6_exp.tex
\input{tables/extreme.tex}

In this section, we perform an extensive empirical evaluation of our methods (\ourbase and \ourmethod) on various GLWI scenarios. 
Our experiments seek to answer the following questions: 

\noindent\textbf{RQ1:} How \textit{effective} are our methods in extreme GLWI scenario?

\noindent\textbf{RQ2:} Can our methods \textit{generally perform well} in various basic GLWI scenarios? 

\noindent\textbf{RQ3:} How \textit{efficient} are our methods in terms of times and space?

\noindent\textbf{RQ4:} How do the key designs and hyper-parameters influence the performance of our methods?

\subsection{Experimental Setups}
\noindent\textbf{Datasets. }
We adopt 8 publicly available real-world graph datasets for evaluation, including Cora~\cite{sen2008collective}, CiteSeer~\cite{sen2008collective}, PubMed~\cite{sen2008collective}, Amazon Photo~\cite{shchur2018pitfalls}, Amazon Computers~\cite{shchur2018pitfalls}, CoAuthor CS~\cite{shchur2018pitfalls}, CoAuthor Physics~\cite{shchur2018pitfalls}, and ogbn-arxiv~\cite{hu2020open}. More details and statistics of datasets are summarized in Appendix~\ref{appendix:dset}.

\noindent\textbf{GLWI scenario implementations. }
We simulate various GLWI scenarios by applying stochastic perturbation on graph data and limiting the number of training nodes. Specifically, to build weak structure, we randomly remove $50\%$ of edges from the original graph structure. To construct weak features, we randomly replace $50\%$ of entries in the feature matrix with $0$. For datasets except ogbn-arxiv, we randomly select $5$ nodes per class to form the training set in weak-label scenario, while this number in other scenarios is $20$. We sample $30$ nodes per class for validation and the rest for testing. For ogbn-arxiv dataset, we randomly select $2\%$ nodes for training with weak labels, and the validation and testing sets follow the official setting~\cite{hu2020open}. In extreme scenario, we construct the insufficient structure, features, and labels via the above strategies respectively and combine them together. 

\noindent\textbf{Baselines. }
We compare our methods with four groups of baselines: 1)~conventional GNNs, including GCN~\cite{kipf2017semi}, GAT~\cite{velivckovic2018graph}, APPNP~\cite{gasteiger2018predict}, and SGC~\cite{wu2019simplifying}; 2)~GNNs with graph structure learning, including Pro-GNN~\cite{jin2020graph}, IDGL~\cite{chen2020iterative}, GEN~\cite{wang2021graph}, and SimP-GCN~\cite{jin2021node}; 3)~GNNs with feature completion, including GINN~\cite{spinelli2020missing} and GCN$_{MF}$~\cite{taguchi2021graph}; 4)~label-efficient GNNs, including M3S~\cite{sun2020multi}, CGPN~\cite{wan2021contrastive}, Meta-PN~\cite{ding2022meta}, and GRAND~\cite{feng2020graph}. In the efficiency analysis, we consider four scalable GNNs, including GraphSAGE~\cite{hamilton2017inductive}, ClusterGCN~\cite{chiang2019cluster}, PPRGo~\cite{bojchevski2020scaling}, and GAMLP~\cite{zhang2022graph}.

\noindent\textbf{Experimental Details. }
For all experiments, we report the averaged test accuracy and standard deviation over 5 trials. 
For our methods, we perform grid search to select the best hyper-parameters on validation set. We also search for optimal hyper-parameters for baseline methods during reproduction. More implementation details are demonstrated in Appendix~\ref{appendix:exp_detail}. Our code is available at \url{https://github.com/yixinliu233/D2PT}.

\input{tables/weak_structure.tex}

\subsection{Performance Comparison (RQ1)}

To \textbf{RQ1}, we compare \ourbase and \ourmethod with $14$ baselines in extreme GLWI scenario, and the results are illustrated in Table~\ref{tab:extreme}. We have the following observations: 1)~\ourmethod achieves state-of-the-art performance on 7 of 8 benchmark datasets and achieves runner-up accuracy on the rest. These results demonstrate the remarkable effectiveness of \ourmethod in learning from graph data with entangled weak information. 2)~Among conventional GNNs, APPNP shows impressive performance on most datasets, which demonstrates the significance of large $s_p$ and graph diffusion mechanism. 3)~The single-aspect GLWI methods have limited improvement compared to conventional GNNs, or even perform worse on some datasets. This observation indicates that only considering the incompleteness in a single aspect may ignore the mutual effect among weak information. 4)~Some single-aspect methods show ``OOM'' when confronting larger datasets, illustrating the heavy computational costs of these methods. 5)~Compared to conventional GNNs, \ourbase achieves superior or comparable performance, which verifies that \ourbase is a strong baseline for GLWI with high efficiency.

\subsection{Generalization Analysis (RQ2)}

\input{tables/weak_feature.tex}

To verify the ability of our methods in handling various basic GLWI scenarios, 
we execute them on graph data with weak structure, features, and labels, and report the results in Table~\ref{tab:weak_structure}, \ref{tab:weak_feature}, and \ref{tab:weak_label}, respectively. From the results, we find that \ourmethod generally outperforms the baseline methods in all scenarios. Moreover, our base model \ourbase also achieves competitive performance, especially on data with weak features. The superior performance illustrates the powerful generalization capability of our proposed methods in learning from graph data with different imperfections.

\subsection{Efficiency Analysis (RQ3)}

\input{tables/weak_label.tex}

\begin{figure}[!t]
 \centering
 \vspace{-0.2cm}
 \subfigure[Accuracy w.r.t. time usage per epoch]{
   \includegraphics[width=0.225\textwidth]{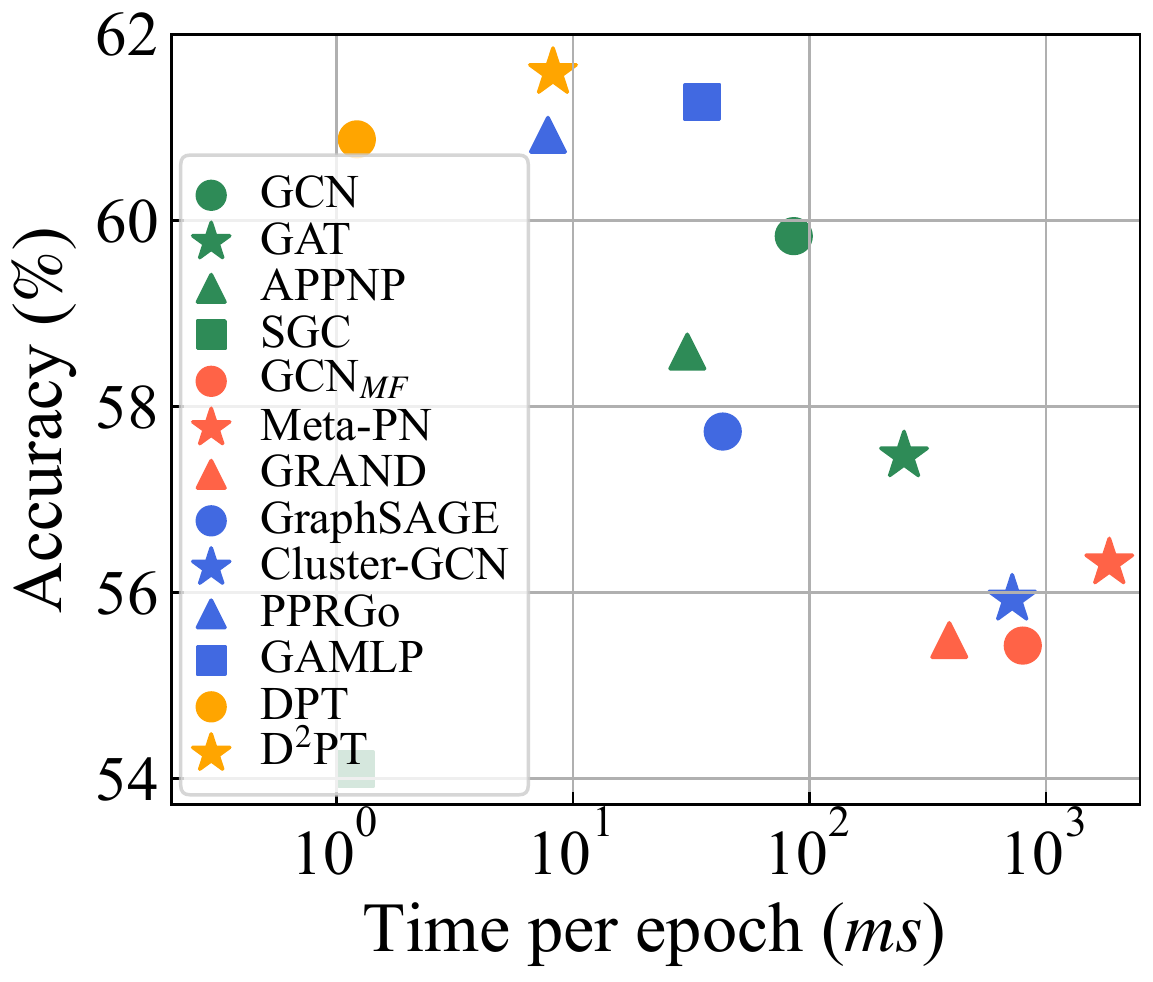}
   \label{subfig:time_eff}
 }
 \subfigure[Accuracy w.r.t. memory usage]{
   \includegraphics[width=0.225\textwidth]{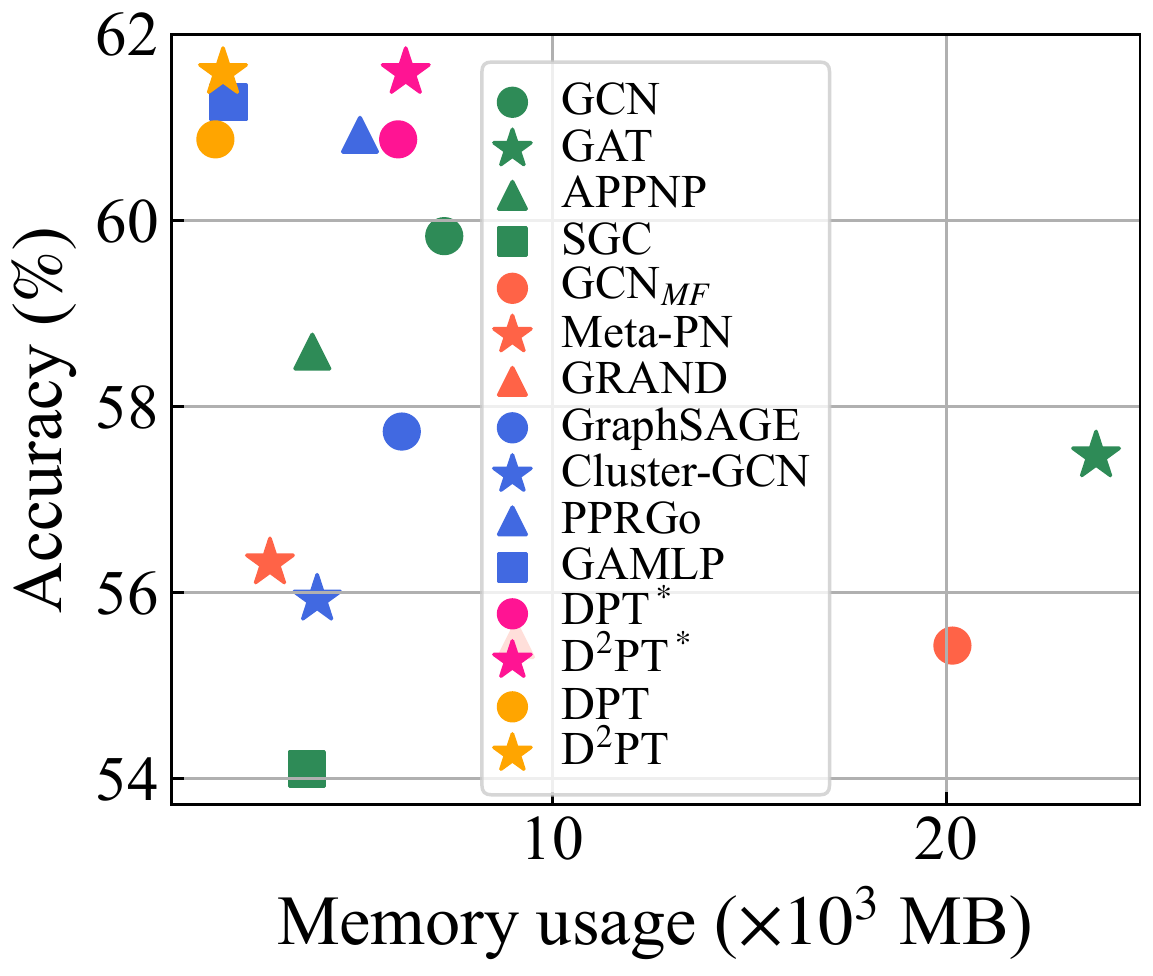}
   \label{subfig:space_eff}
 } 
 \vspace{-0.4cm}
 \caption{Efficiency analysis on ogbn-arxiv dataset.}
 \label{fig:efficiency}
 \vspace{-0.5cm}
\end{figure}

We analyze the efficiency of \ourbase and \ourmethod on ogbn-arxiv dataset in terms of running time per epoch and memory usage on graphic cards. The results are shown in Fig.~\ref{fig:efficiency} where DPT$^*$ and D$^2$PT$^*$ indicate the models that save full data into graphic cards for efficient validation and testing. From Fig.~\ref{subfig:time_eff}, we find that \ourmethod enjoys high running efficiency and state-of-the-art performance compared to baselines. For instance, \ourmethod is 3.69x faster than APPNP, 30x faster than GAT, and 225x faster than Meta-PN. Meanwhile, \ourbase has extremely high efficiency (close to SGC) while still achieving excellent performance. From the perspective of memory usage, thanks to the adjacency decoupling and mini-batch semi-supervised learning designs, the memory usages of \ourmethod and \ourbase are smaller than 2000MB, which verifies their space efficiency. Even if we load the full dataset for efficient evaluation, their memory usages are still comparable to most baselines. An interesting finding is that two large-$s_p$ scalable GNNs, PPRGo and GAMLP, also yield competitive performance, illustrating the advantage of long-range propagation in GLWI scenarios.

\subsection{Ablation and Parameter Studies (RQ4)}

\noindent\textbf{Effect of key components.} 
We illustrate the effect of dual-channel training and contrastive prototype alignment by removing the corresponding losses. As is shown in the middle block of Table~\ref{tab:ablation}, both of these designs give significant performance improvement over the base model, and the contrastive prototype alignment seems to contribute more. Moreover, \ourmethod, which jointly considers both of them, produces the best results. 

\noindent\textbf{Selection of global graph.} 
Besides generating kNN graph from $\overline{\mathbf{X}}$, we also attempt two alternative strategies: generating kNN graph from raw features (kNN from $\mathbf{X}$) and directly using raw features as the augmented data (Aug. w/o prop). From Table~\ref{tab:ablation}, we can observe that these strategies produce poor performance. The observation demonstrates the high quality of the kNN graph from $\overline{\mathbf{X}}$ where features are completed by long-range propagation. 

\input{tables/ablation.tex}

\begin{figure}[!t]
 \centering
 \vspace{-0.1cm}
 \subfigure[Accuracy w.r.t. $T$]{
   \includegraphics[height=0.17\textwidth]{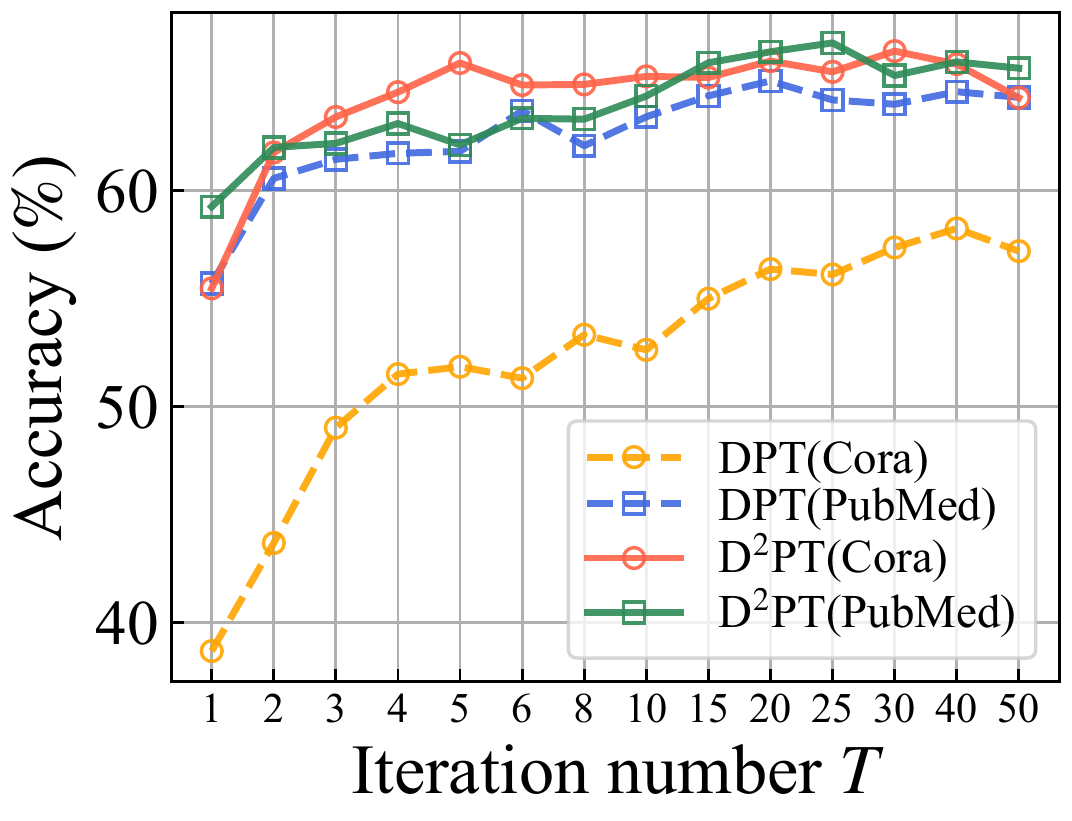}
   \label{subfig:param_t}
 }
 \subfigure[t-SNE visualization]{
   \includegraphics[height=0.17\textwidth]{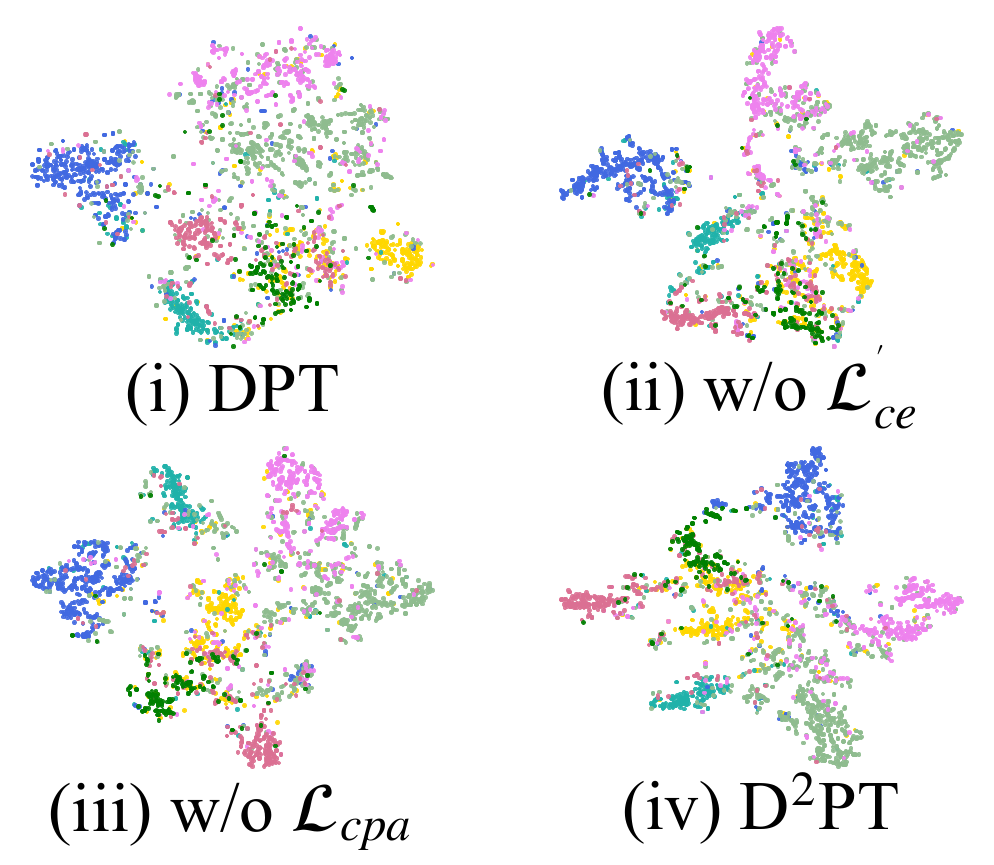}
   \label{subfig:tsne}
 } 
 \vspace{-0.4cm}
 \caption{Parameter study and visualization results.}
 \label{fig:param}
 \vspace{-0.3cm}
\end{figure}

\noindent\textbf{Effect of propagation step.} 
To study the impact of propagation step $s_p$, we tune the iteration number $T$ ($T=s_p$ in our methods) from $1$ to $50$ on two datasets. As shown in Fig.~\ref{subfig:param_t}, the performance of \ourbase and \ourmethod generally increases with the growth of $T$. The results verify our statement that GNNs with long-range propagation tend to perform better in GLWI scenarios. Also, we can find that the accuracy becomes stable when $T\geq20$, indicating that a moderate propagation step can provide sufficient information for GLWI.

\noindent\textbf{Visualization.} 
For qualitative analysis, we provide visualizations of the learned representations $\mathbf{H}$ via t-SNE~\cite{van2008visualizing}. The results on Cora dataset are presented in Fig.\ref{subfig:tsne} where nodes in the same color are from the same class. We can observe that the result from \ourbase, where nodes with different labels are mixed together, is not satisfactory enough. The possible reason is that the embeddings of stray nodes are less distinguishable in the latent space. By introducing the global graph with $\mathcal{L}'_{ce}$ or $\mathcal{L}_{cpa}$, the decision boundary becomes clearer in (ii) and (iii). With the guidance of two auxiliary losses, \ourmethod performs best, proved by the more compact structure and distinct boundary of the learned representations in (iv). 

%% file: tables/extreme.tex
\begin{table*}[t]
\vspace{-0.1cm}
\centering
\caption{Results in terms of classification accuracy (in percent $\pm$ standard deviation) in extreme GLWI scenario. OOM indicates Out-Of-Memory on a 24GB GPU. The best and runner-up results are highlighted with \textcolor{sotacolor}{\textbf{bold}} and \textcolor{runupcolor}{\underline{underline}}, respectively.} 
\label{tab:extreme}
\vspace{-0.3cm}
\resizebox{0.9\textwidth}{!}{
 \begin{tabular}{p{1.6cm}|p{1.6cm}<{\centering} p{1.6cm}<{\centering} p{1.6cm}<{\centering} p{1.6cm}<{\centering} p{1.6cm}<{\centering} p{1.6cm}<{\centering} p{1.6cm}<{\centering} p{1.6cm}<{\centering}}
\toprule
Methods & Cora & CiteSeer & PubMed  & Amz. Photo & Amz. Comp.  & Co. CS & Co. Physics & ogbn-arxiv\\
\midrule
GCN & $46.80{\scriptstyle\pm3.20}$ & $42.31{\scriptstyle\pm3.60}$ & $62.30{\scriptstyle\pm2.82}$ & $83.44{\scriptstyle\pm1.61}$ & $71.22{\scriptstyle\pm1.24}$ & $84.14{\scriptstyle\pm1.40}$ & $88.83{\scriptstyle\pm1.44}$ & $59.83{\scriptstyle\pm0.31}$  \\
GAT & $48.79{\scriptstyle\pm5.89}$ & $43.86{\scriptstyle\pm4.43}$ & $63.07{\scriptstyle\pm2.82}$ & $82.33{\scriptstyle\pm1.00}$ & $71.81{\scriptstyle\pm4.51}$ & $82.30{\scriptstyle\pm2.01}$ & $88.53{\scriptstyle\pm1.65}$ & $57.47{\scriptstyle\pm0.31}$  \\
APPNP & $54.91{\scriptstyle\pm5.18}$ & $43.59{\scriptstyle\pm5.07}$ & $64.77{\scriptstyle\pm3.56}$ & $83.78{\scriptstyle\pm1.08}$ & $\runup{73.46{\scriptstyle\pm2.84}}$ & $84.59{\scriptstyle\pm1.73}$ & $90.22{\scriptstyle\pm0.99}$ & $58.59{\scriptstyle\pm1.02}$  \\
SGC & $48.06{\scriptstyle\pm6.00}$ & $40.56{\scriptstyle\pm3.51}$ & $61.89{\scriptstyle\pm3.30}$ & $82.12{\scriptstyle\pm1.26}$ & $71.36{\scriptstyle\pm2.12}$ & $83.82{\scriptstyle\pm1.50}$ & $88.20{\scriptstyle\pm1.22}$ & $54.10{\scriptstyle\pm0.73}$  \\
\midrule
Pro-GNN & $53.28{\scriptstyle\pm7.27}$ & $44.92{\scriptstyle\pm1.62}$ & OOM & OOM & OOM & OOM & OOM & OOM  \\
IDGL & $46.95{\scriptstyle\pm4.60}$ & $45.98{\scriptstyle\pm2.64}$ & $63.35{\scriptstyle\pm4.19}$ & $\runup{83.85{\scriptstyle\pm1.16}}$ & $73.09{\scriptstyle\pm1.94}$ & $83.73{\scriptstyle\pm1.97}$ & OOM & OOM  \\
GEN & $52.72{\scriptstyle\pm3.82}$ & $49.49{\scriptstyle\pm3.70}$ & OOM & OOM & OOM & OOM & OOM & OOM  \\
SimP-GCN & $48.72{\scriptstyle\pm3.89}$ & $48.31{\scriptstyle\pm4.45}$ & $64.25{\scriptstyle\pm3.80}$ & $82.94{\scriptstyle\pm1.84}$ & $71.17{\scriptstyle\pm2.06}$ & $\sota{87.13{\scriptstyle\pm1.30}}$ & $89.72{\scriptstyle\pm1.43}$ & OOM  \\
\midrule
GINN & $47.07{\scriptstyle\pm3.80}$ & $41.82{\scriptstyle\pm2.09}$ & $61.80{\scriptstyle\pm2.97}$ & $82.09{\scriptstyle\pm1.21}$ & $70.33{\scriptstyle\pm1.09}$ & $81.78{\scriptstyle\pm1.25}$ & OOM & OOM  \\
GCN$_{MF}$ & $47.13{\scriptstyle\pm2.74}$ & $43.64{\scriptstyle\pm4.72}$ & $61.91{\scriptstyle\pm4.05}$ & $83.22{\scriptstyle\pm0.98}$ & $73.17{\scriptstyle\pm2.11}$ & $83.49{\scriptstyle\pm1.88}$ & $88.56{\scriptstyle\pm1.56}$ & $55.43{\scriptstyle\pm0.66}$  \\
\midrule
M3S & $56.15{\scriptstyle\pm3.89}$ & $52.13{\scriptstyle\pm5.03}$ & $63.65{\scriptstyle\pm5.01}$ & $82.64{\scriptstyle\pm2.96}$ & $71.96{\scriptstyle\pm1.49}$ & $83.25{\scriptstyle\pm2.07}$ & $88.99{\scriptstyle\pm3.04}$ & OOM \\
CGPN & $59.29{\scriptstyle\pm1.54}$ & $\runup{54.62{\scriptstyle\pm2.74}}$ & $64.47{\scriptstyle\pm1.17}$ & $82.60{\scriptstyle\pm2.36}$ & $72.36{\scriptstyle\pm2.70}$ & $82.92{\scriptstyle\pm2.54}$ & OOM & OOM  \\
Meta-PN & $53.05{\scriptstyle\pm5.97}$ & $45.60{\scriptstyle\pm5.84}$ & ${64.88{\scriptstyle\pm4.02}}$ & $82.46{\scriptstyle\pm1.91}$ & $72.19{\scriptstyle\pm3.11}$ & $83.95{\scriptstyle\pm2.43}$ & $89.28{\scriptstyle\pm1.01}$ & $56.32{\scriptstyle\pm0.86}$  \\
GRAND & $\runup{63.38{\scriptstyle\pm1.40}}$ & $46.23{\scriptstyle\pm4.44}$ & $63.97{\scriptstyle\pm3.82}$ & $82.66{\scriptstyle\pm4.85}$ & $71.84{\scriptstyle\pm5.20}$ & $84.93{\scriptstyle\pm1.50}$ & $90.33{\scriptstyle\pm0.67}$ & $55.49{\scriptstyle\pm0.47}$  \\
\midrule
\ourbase & $56.37{\scriptstyle\pm5.97}$ & $46.06{\scriptstyle\pm4.56}$ & $\runup{65.08{\scriptstyle\pm3.13}}$ & $82.84{\scriptstyle\pm1.46}$ & $73.13{\scriptstyle\pm1.70}$ & $84.32{\scriptstyle\pm2.10}$ & $\runup{90.42{\scriptstyle\pm1.13}}$ & $\runup{60.87{\scriptstyle\pm0.18}}$  \\
\ourmethod & $\sota{66.00{\scriptstyle\pm1.20}}$ & $\sota{56.99{\scriptstyle\pm2.23}}$ & $\sota{66.43{\scriptstyle\pm2.45}}$ & $\sota{84.12{\scriptstyle\pm1.66}}$ & $\sota{74.96{\scriptstyle\pm1.79}}$ & $\runup{85.62{\scriptstyle\pm1.75}}$ & $\sota{91.41{\scriptstyle\pm0.89}}$ & $\sota{61.59{\scriptstyle\pm0.59}}$  \\
\bottomrule
\end{tabular}}
\vspace{-0.2cm}
\end{table*}

%% file: tables/weak_structure.tex
\begin{table}[t]
\centering
\vspace{-0.1cm}
\caption{Results in terms of classification accuracy in graph learning with weak structure.} 
\label{tab:weak_structure}
\vspace{-0.3cm}
\resizebox{0.9\columnwidth}{!}{
\begin{tabular}{p{1.8cm}|p{1.6cm}<{\centering} p{1.6cm}<{\centering} p{1.6cm}<{\centering}}
\toprule
Methods & Cora & CiteSeer & PubMed  \\
\midrule
GCN & $74.07{\scriptstyle\pm2.22}$ & $61.83{\scriptstyle\pm1.18}$ & $74.55{\scriptstyle\pm2.02}$  \\
GAT & $73.03{\scriptstyle\pm1.37}$ & $61.55{\scriptstyle\pm2.07}$ & $73.89{\scriptstyle\pm1.80}$  \\
APPNP & $\runup{76.81{\scriptstyle\pm1.16}}$ & $63.49{\scriptstyle\pm1.47}$ & $75.85{\scriptstyle\pm0.65}$  \\
SGC & $75.34{\scriptstyle\pm0.48}$ & $64.55{\scriptstyle\pm1.62}$ & $73.06{\scriptstyle\pm1.64}$  \\
\midrule
Pro-GNN & $75.72{\scriptstyle\pm1.48}$ & $64.57{\scriptstyle\pm1.68}$ & OOM  \\
IDGL & $76.17{\scriptstyle\pm2.00}$ & $64.58{\scriptstyle\pm1.54}$ & $\runup{76.19{\scriptstyle\pm1.96}}$  \\
GEN & $76.34{\scriptstyle\pm0.88}$ & $65.34{\scriptstyle\pm2.05}$ & OOM  \\
SimP-GCN & $76.42{\scriptstyle\pm1.85}$ & $\runup{67.08{\scriptstyle\pm2.10}}$ & $75.82{\scriptstyle\pm0.94}$  \\
\midrule
\ourbase & $76.05{\scriptstyle\pm1.31}$ & $64.03{\scriptstyle\pm2.15}$ & ${75.22{\scriptstyle\pm1.57}}$  \\
\ourmethod & $\sota{77.92{\scriptstyle\pm1.53}}$ & $\sota{67.22{\scriptstyle\pm1.35}}$ & $\sota{76.42{\scriptstyle\pm0.97}}$  \\
\bottomrule
\end{tabular}}
\vspace{-0.5cm}
\end{table}

%% file: tables/weak_feature.tex
\begin{table}[t]
\centering
\vspace{-0.1cm}
\caption{Results in terms of classification accuracy in graph learning with weak features.} 
\label{tab:weak_feature}
\vspace{-0.3cm}
\resizebox{0.9\columnwidth}{!}{
\begin{tabular}{p{1.8cm}|p{1.6cm}<{\centering} p{1.6cm}<{\centering} p{1.6cm}<{\centering}}
\toprule
Methods & Cora & CiteSeer & PubMed  \\
\midrule
GCN & $76.40{\scriptstyle\pm0.67}$ & $62.44{\scriptstyle\pm3.06}$ & $75.42{\scriptstyle\pm1.16}$  \\
GAT & $77.26{\scriptstyle\pm1.18}$ & $62.07{\scriptstyle\pm3.41}$ & $75.48{\scriptstyle\pm0.80}$  \\
APPNP & ${79.46{\scriptstyle\pm1.13}}$ & $64.68{\scriptstyle\pm2.01}$ & ${77.66{\scriptstyle\pm0.68}}$  \\
SGC & $78.74{\scriptstyle\pm1.56}$ & $64.90{\scriptstyle\pm2.74}$ & $72.47{\scriptstyle\pm1.75}$  \\
\midrule
GINN & $78.04{\scriptstyle\pm1.22}$ & $64.11{\scriptstyle\pm1.44}$ & $75.29{\scriptstyle\pm0.89}$  \\
GCN$_{MF}$ & $79.03{\scriptstyle\pm0.99}$ & ${65.23{\scriptstyle\pm2.39}}$ & $\runup{76.14{\scriptstyle\pm1.55}}$  \\
\midrule
\ourbase & $\runup{79.70{\scriptstyle\pm0.82}}$ & $\runup{65.50{\scriptstyle\pm2.29}}$ & ${75.87{\scriptstyle\pm1.09}}$  \\
\ourmethod & $\sota{80.66{\scriptstyle\pm0.53}}$ & $\sota{66.42{\scriptstyle\pm2.14}}$ & $\sota{78.26{\scriptstyle\pm0.84}}$  \\
\bottomrule
\end{tabular}}
\vspace{-0.5cm}
\end{table}

%% file: tables/weak_label.tex
\begin{table}[t]
\centering
\vspace{-0.1cm}
\caption{Results in terms of classification accuracy in graph learning with weak labels.} 
\label{tab:weak_label}
\vspace{-0.3cm}
\resizebox{0.9\columnwidth}{!}{
\begin{tabular}{p{1.8cm}|p{1.6cm}<{\centering} p{1.6cm}<{\centering} p{1.6cm}<{\centering}}
\toprule
Methods & Cora & CiteSeer & PubMed  \\
\midrule
GAT & $74.39{\scriptstyle\pm1.97}$ & $57.62{\scriptstyle\pm3.69}$ & $68.98{\scriptstyle\pm3.39}$  \\
APPNP & ${78.08{\scriptstyle\pm2.39}}$ & $56.20{\scriptstyle\pm5.33}$ & $71.30{\scriptstyle\pm3.11}$  \\
SGC & $76.28{\scriptstyle\pm2.13}$ & $57.99{\scriptstyle\pm3.90}$ & $66.45{\scriptstyle\pm4.69}$  \\
\midrule
M3S & $76.30{\scriptstyle\pm3.08}$ & $64.17{\scriptstyle\pm2.43}$ & $69.45{\scriptstyle\pm5.48}$  \\
CGPN & $76.09{\scriptstyle\pm1.75}$ & $\runup{64.84{\scriptstyle\pm2.56}}$ & $69.94{\scriptstyle\pm3.25}$  \\
Meta-PN & $77.83{\scriptstyle\pm1.76}$ & $61.86{\scriptstyle\pm3.82}$ & $\runup{71.96{\scriptstyle\pm3.46}}$  \\
GRAND & $\runup{78.09{\scriptstyle\pm1.96}}$ & $59.52{\scriptstyle\pm4.05}$ & $69.17{\scriptstyle\pm3.53}$  \\
\midrule
\ourbase & ${77.55{\scriptstyle\pm2.40}}$ & $60.11{\scriptstyle\pm2.30}$ & $70.07{\scriptstyle\pm2.40}$  \\
\ourmethod & $\sota{79.00{\scriptstyle\pm2.05}}$ & $\sota{67.94{\scriptstyle\pm1.14}}$ & $\sota{72.17{\scriptstyle\pm1.79}}$  \\
\bottomrule
\end{tabular}}
\vspace{-0.4cm}
\end{table}

%% file: tables/ablation.tex
\begin{table}[t]
\centering
\vspace{-0.1cm}
\caption{Performance of \ourmethod and its variants.} 
\label{tab:ablation}
\vspace{-0.3cm}
\resizebox{0.9\columnwidth}{!}{
\begin{tabular}{p{1.8cm}|p{1.6cm}<{\centering} p{1.6cm}<{\centering} p{1.6cm}<{\centering}}
\toprule
Methods & Cora & CiteSeer & PubMed  \\
\midrule
\ourmethod & $66.00{\scriptstyle\pm1.20}$ & $56.99{\scriptstyle\pm2.23}$ & $66.43{\scriptstyle\pm2.45}$  \\
\midrule
w/o $\mathcal{L}'_{ce}$ & $61.97{\scriptstyle\pm5.22}$ & $53.56{\scriptstyle\pm4.04}$ & $64.33{\scriptstyle\pm2.87}$  \\
w/o $\mathcal{L}_{cpa}$ & $61.00{\scriptstyle\pm1.77}$ & $48.37{\scriptstyle\pm2.53}$ & $63.97{\scriptstyle\pm5.18}$  \\
Base (\ourbase) & $56.37{\scriptstyle\pm5.97}$ & $46.06{\scriptstyle\pm4.56}$ & $65.08{\scriptstyle\pm3.13}$  \\
\midrule
kNN from $\mathbf{X}$ & $54.52{\scriptstyle\pm3.22}$ & $40.56{\scriptstyle\pm3.01}$ & $63.29{\scriptstyle\pm4.78}$  \\
Aug. w/o prop & $52.68{\scriptstyle\pm5.51}$ & $41.22{\scriptstyle\pm4.04}$ & $64.56{\scriptstyle\pm3.51}$  \\
\bottomrule
\end{tabular}}
\vspace{-0.3cm}
\end{table}

%% file: 7_conclu.tex
In this paper, we make the first attempt towards graph learning with weak information (GLWI), a practical yet challenging learning problem on graph-structured data with incomplete structure, features, and labels. 
With discussion and empirical analysis, we expose the key to addressing GLWI problem is effective information propagation, and figure out two crucial criteria for model design: enabling long-range propagation and handling stray nodes. 
Following these criteria, we propose a novel GNN model termed \ourmethod that enjoys high efficiency for long-range propagation and solves stray node problem with an augmented global graph and dual-channel architecture. 
Extensive experiments demonstrate the effectiveness of \ourmethod in multiple GLWI scenarios. 
In the future, promising follow-up directions include: 1) exploring GLWI for data with more challenging data deficiencies, such as noisy features, noisy edges, and imbalanced label distribution; 2) applying GLWI to more downstream tasks, e.g., graph classification and link prediction; and 3) unsupervised graph learning for incomplete data.

%% file: 8_appendix.tex
\section{Details of Motivated Experiments} \label{appendix:moti_exp}

\subsection{Details of Propagation Step Experiment} \label{appendix:sp_exp}

In this experiment, we implement 5 mainstream GNNs for comparison, including GCN~\cite{kipf2017semi}, GAT~\cite{velivckovic2018graph}, SGC~\cite{wu2019simplifying}, PPNP~\cite{gasteiger2018predict}, and APPNP~\cite{gasteiger2018predict}. Note that we do not consider deeper GCN~\cite{kipf2017semi} and GAT~\cite{velivckovic2018graph} because they are proven to be ineffective with more layers~\cite{kipf2017semi, zhang2022model}. The source codes for reproduction are from PyTorch Geometric examples\footnote{\url{https://github.com/pyg-team/pytorch_geometric/tree/master/benchmark/citation}} and original releases\footnote{\url{https://github.com/gasteigerjo/ppnp}}. We run 5 trials for each method and report the average accuracy. For all methods, we find the best hyper-parameters using grid search with the following search space:

\begin{itemize}
    \item Number of layers/propagation iterations: \{1, 2\} (for small-$s_p$ models); \{3, 5, 8, 10\} (for large-$s_p$ models)
    \item Hidden unit: \{32, 64, 128, 256\}
    \item Learning rate: \{1e-1, 1e-2, 1e-3\}
    \item Number of epochs: \{100, 200, 500\}
    \item Number of attention heads (GAT): \{4, 8, 16\}
    \item Restart probability $\alpha$ (PPNP/APPNP): \{0.01, 0.05, 0.1, 0.2\}
\end{itemize}

\subsection{Details of Weak Information Scenario Combination Experiment} \label{appendix:comb_exp}

To eliminate the influence of incomplete degrees of structure, features, and labels, in this experiment, we tune the missing rates to ensure our model has similar performance on three basic scenarios. In concrete, we first acquire the accuracy $acc_{wl}$ in weak label sencario where $5$ samples are used for training per class. Then, we search the missing rates of features and edges from $0.05$ to $0.95$ and define the settings where accuracy is closest to $acc_{wl}$ as the weak feature and weak structure scenarios. The specific settings are given in Table~\ref{tab:comb_setting}. 
We conduct this experiment on APPNP~\cite{gasteiger2018predict} model, where learning rate is $0.01$, hidden unit is $64$, propagation iteration is $20$, and restart probability $\alpha$ is $0.05$. We repeat the experiment for 5 times and report the average values and standard deviations in the bar chart. 

\subsection{Details of Stray Node Experiments} \label{appendix:stray_exp}

In these experiments, we simulate the extreme GLWI scenario by setting the edge missing rate to $0.5$, feature missing rate to $0.5$, number of training samples per class to $5$. 

For the feature completion experiment, we consider an SGC-like~\cite{wu2019simplifying} propagation strategy, i.e., multiplying the feature matrix $\mathbf{X}$ by normalized adjacency matrix $\tilde{\mathbf{A}}$ for $s_p$ times, where $s_p$ is the number of propagation iteration. Then, we compute the L2 distance between the original feature vectors and propagated feature vectors of all the nodes. Here we set $s_p=20$ to ensure adequate propagation. The average L2 distance of 5 runs of experiments is reported. 

For the supervision signal spreading experiment, we acquire the testing accuracy with an APPNP model where learning rate is $0.01$, hidden unit is $64$, propagation iteration is $20$, and restart probability $\alpha$ is $0.05$. The average accuracy of 5 runs of experiments is reported.

\section{Discussion for \ourbase} \label{appendix:dpt_discuss}

\begin{figure}[!t]
\centering
  \includegraphics[width=0.8\columnwidth]{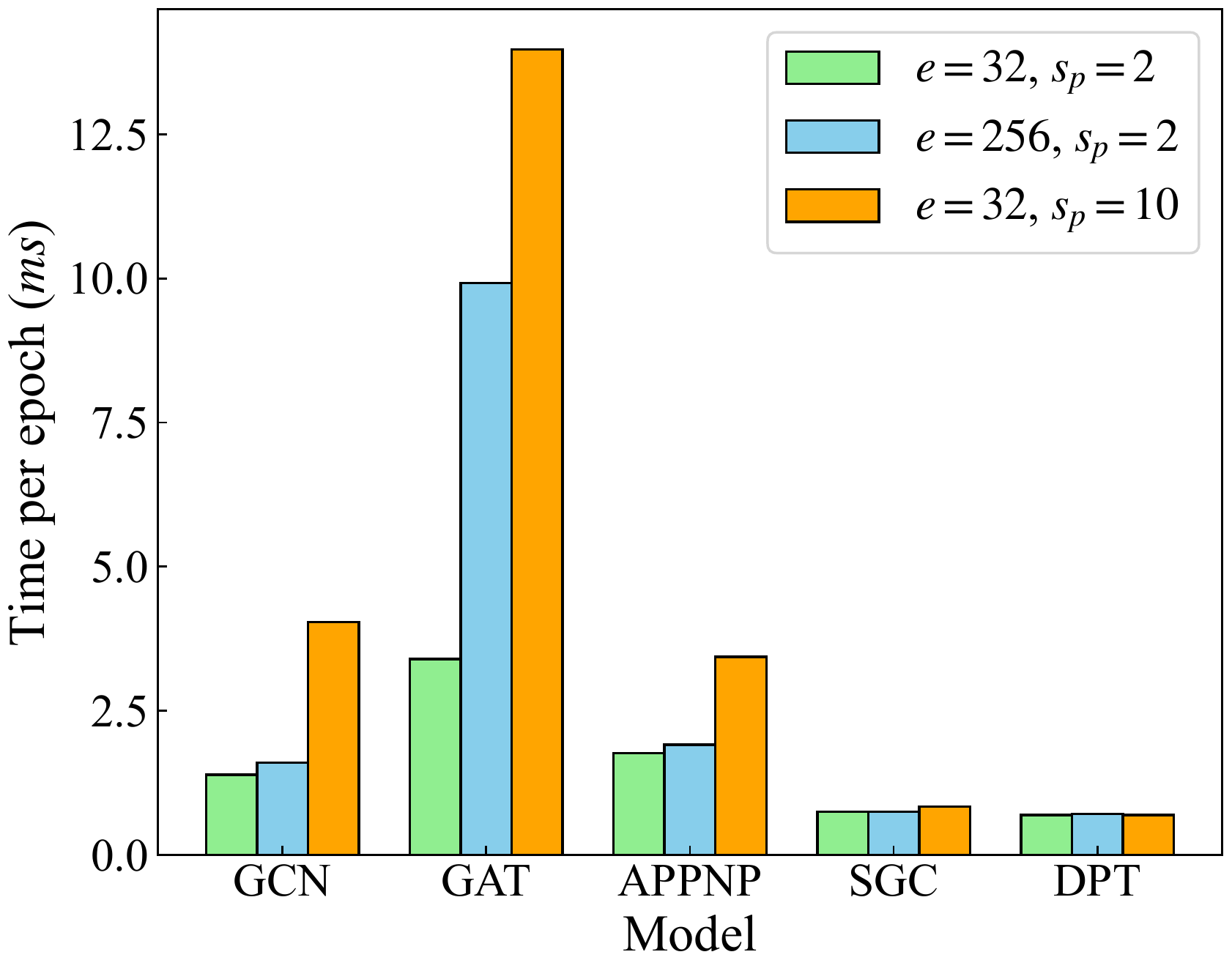}
  \vspace{-0.3cm}
  \caption{Comparison of training time.
  }
  \label{subfig:basemodel_speed}
  \vspace{-0.3cm}
\end{figure}

In this section, we discuss the superior properties of \ourbase in the following paragraphs. 
A comparison between \ourbase and conventional GNNs is illustrated in Table~\ref{tab:base_model_compare}. 
Meanwhile, on three settings with different hidden unit $e$ and propagation step $s_p$, we conduct an experimental comparison w.r.t. running time per epoch in Fig.~\ref{subfig:basemodel_speed}. 

\noindent\textbf{Only propagation once.} 
Different from \texttt{TPTP} models and \texttt{TTPP} models where propagation operations are executed at each training iteration, \ourbase only propagates {once} during the whole training process. Since the propagation phase (Eq.~\ref{eq:prop}) is parameter-free and directly based on raw features, the computation of $\overline{\mathbf{X}}$ can be done in a preprocessing step. In this case, the time complexity of model training is decoupled from $s_p$, leading to higher efficiency when $s_p$ is large. As shown in Table~\ref{tab:base_model_compare}, the training time of \ourbase is independent to $s_p$. In Fig.~\ref{subfig:basemodel_speed}, we can find that the time complexities of GCN~\cite{kipf2017semi}, GAT~\cite{velivckovic2018graph}, and APPNP~\cite{gasteiger2018predict} significantly increase when $s_p$ gets larger. To sum up, \ourbase can enable long-range propagation while preserving high efficiency.

\noindent\textbf{Decoupled from adjacency matrix.} 
After acquiring $\overline{\mathbf{X}}$ by the one-time propagation, we only need to train the MLP-based transformation  model with $\overline{\mathbf{X}}$ as its input. In other words, the training process is decoupled from adjacency matrix, and then all the samples (i.e., row vectors) in $\overline{\mathbf{X}}$ can be loaded independently. With this property, we can train the model in a flexible and efficient way. For graphs with billions of nodes, we can use partition-based propagation~\cite{chiang2019cluster, zeng2020graphsaint} to generate $\overline{\mathbf{X}}$, and then use mini-batch strategy to train the transformation model. In this way, \ourbase can be applied to large-scale graphs without exponential memory cost. Moreover, in semi-supervised scenarios, we can only train the model on the rows of $\overline{\mathbf{X}}$ that belong to training samples, avoiding the high computational cost of running models on all nodes. Thanks to this merit, in Fig.~\ref{subfig:basemodel_speed}, \ourbase achieves faster training speed than most GNNs. 

\begin{table}[t]
\centering
\caption{Detail settings of scenario combination experiment.}
\label{tab:comb_setting}
\vspace{-0.2cm}
\resizebox{1\columnwidth}{!}{
\begin{tabular}{l | ccc}
\toprule
Dataset & \tabincell{c}{Edge missing rate\\ of ``WS'' scenario}  & \tabincell{c}{Feature missing rate\\ of ``WF'' scenario} & \tabincell{c}{\#train samples per class\\ of ``WL'' scenario}  \\
\hline
Cora & $0.35$ & $0.7$ & $5$ \\
CiteSeer & $0.9$ & $0.85$ & $5$ \\
PubMed & $0.85$ & $0.85$ & $5$  \\
\bottomrule
\end{tabular}}
\end{table}

\begin{table}[t]
\centering
\caption{Comparison between \ourbase and GNNs. ``Time Comp.'' and ``Dec. Adj.'' indicate ``Time complexity'' and ``Decoupled from adjacency matrix during training'', respectively.}
\label{tab:base_model_compare}
\vspace{-0.2cm}
\resizebox{1\columnwidth}{!}{
\begin{threeparttable}
\begin{tabular}{l | cccc}
\toprule
Model & Type  & Time Comp.\tnote{1,2,3}& Dec. Adj. & Diffusion \\
\hline
GCN~\cite{kipf2017semi} & \texttt{PTPT} & $\mathcal{O}(sme + sne^2)$   & - & - \\
GAT~\cite{velivckovic2018graph} & \texttt{PTPT} & $\mathcal{O}(sme + sne^2)$  & - & - \\
SGC~\cite{wu2019simplifying} & \texttt{PPTT} & $\mathcal{O}(s_tn_Le^2)$ & \checkmark & - \\
APPNP~\cite{gasteiger2018predict} & \texttt{TTPP} & $\mathcal{O}(s_pme + s_tne^2)$ & - & \checkmark \\
\midrule
\ourbase & \texttt{PPTT} & $\mathcal{O}(s_tn_Le^2)$ & \checkmark & \checkmark \\
\bottomrule
\end{tabular}
\begin{tablenotes}   
\footnotesize   
\item[1] $s=s_p=s_t$ is the layer number of GCN/GAT. 
\item[2] We omit the complexity related to attention mechanism in GAT. 
\item[3] We omit the complexity at the first transformation layer (related to $d$) for simplify. 
\end{tablenotes} 
\end{threeparttable}
}
\end{table}

\noindent\textbf{Flexible diffusion-based propagation.} 
Recall those models with graph diffusion perform better as we discussed in Sec.~\ref{subsec:large_sp}. in \ourbase, we also introduce this mechanism. To balance efficiency and effectiveness, we use topic-sensitive PageRank~\cite{zeng2020graphsaint} as the diffusion, which has several advantages. Compare to models without diffusion (e.g., SGC~\cite{wu2019simplifying}), \ourbase has a residual connection to add the original features into the output at each propagation step, which emphasizes the ego information and alleviates the over-smoothing issue caused by large $s_p$~\cite{li2019deepgcns,chen2020simple}. Hence, in Fig.~\ref{subfig:wf_propstep}-\ref{subfig:ws_propstep}, \ourbase generally outperforms SGC and has competitive performance compared to other diffusion-based models. Compared to computing diffusion via closed-form solution (e.g., PPNP~\cite{gasteiger2018predict}), our model is more efficient on large-scale graphs. 

\section{Details of Algorithm} \label{appendix:full_algo}

\subsection{Glocal kNN Approximation} \label{appendix:fast_knn}

\begin{figure}[!t]
 \centering
 \subfigure[Adjacency matrix by locality-sensitive kNN approximation.]{
   \includegraphics[height=0.31\columnwidth]{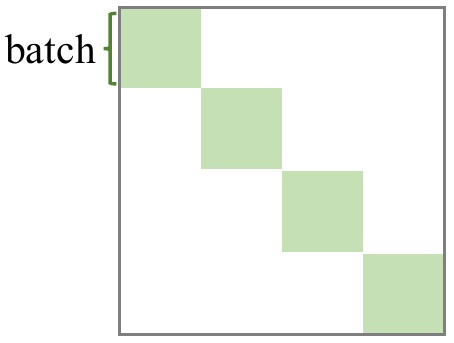}
   \label{subfig:local}
 }  \hspace{1mm}
 \subfigure[Adjacency matrix by glocal kNN approximation.]{
   \includegraphics[height=0.31\columnwidth]{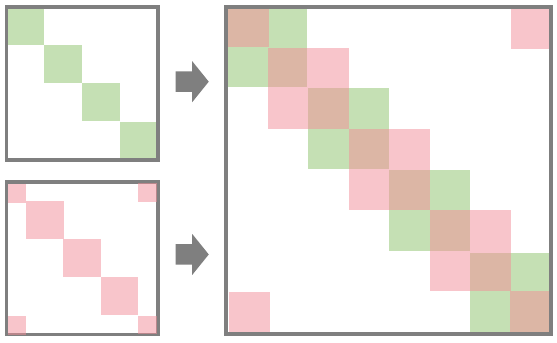}
   \label{subfig:glocal}
 }
 \vspace{-0.2cm}
 \caption{Sketch maps to illustrate kNN approximation.}
 \vspace{-0.3cm}
 \label{fig:glocal}
\end{figure}

Computing a kNN graph often needs a time complexity of $\mathcal{O}(n^2d)$, resulting in the heavy computational cost when $n$ (number of nodes) is overlarge. To tackle this problem, recent efforts~\cite{fatemi2021slaps,halcrow2020grale} introduce a locality-sensitive approximation algorithm for kNN. In this algorithm, the top-k neighbors are selected from a batch of nodes instead of all nodes, which reduces the time complexity to $\mathcal{O}(nbd)$ where $b$ is the batch size for kNN approximation. 

Despite its efficiency, the locality-sensitive approximation algorithm may generate global graph that are not in line with our expectations. As shown in Fig.~\ref{subfig:local}, since the neighbors are estimated from a small batch of nodes, the generated graph is composed by several small fractions with size $b$ rather than a large connected component. Such a situation violates our original intention, i.e., constructing a connected global graph. To fill the gap, we modify the locality-sensitive approximation into a glocal approximation for kNN. In specific, we execute the locality-sensitive approximation algorithm \textit{twice} with different batch splitting. Here, the numbers of neighbors are set to $k_1$ and $k_2$ (ensuring $k_1+k_2=k$), respectively. Then, we combine two generated local kNN graphs into a global kNN graph, as illustrated in Fig.~\ref{subfig:glocal}. In this case, in the generated global graph, each node is potentially connected (with several hops) to all other nodes rather than the nodes within same batch. That is to say, the generated graph can be a large connected graph. At the same time, the complexity of glocal approximation is still $\mathcal{O}(nbd)$, preserving the efficiency. 

\subsection{Algorithm} \label{appendix:algo}

\begin{algorithm}[t]
\caption{The training algorithm of \ourmethod}
\label{alg:overall}
\LinesNumbered
\KwIn{Feature matrix $\mathbf{X}$; Adjacency matrix $\mathbf{A}$; Training label matrix $\mathbf{Y}_L$.} 
\Parameter{Number of epoch $E$; Propagation iteration $T$; Restart Probability $\alpha$; Number of neighbor $k$; Temperature $\tau$; Trade-off coefficients $\gamma_1, \gamma_2$}
\tcc{Pre-processing}
Calculate $\overline{\mathbf{X}} \gets \operatorname{propagation}(\mathbf{X}, \mathbf{A}; T, \alpha)$ via Eq.~(\ref{eq:prop})\\
\eIf{Use glocal kNN}{Calculate $\mathbf{A}' \gets \operatorname{glocal-kNN}(\overline{\mathbf{X}}; k)$ via Appendix~\ref{appendix:fast_knn}}{Calculate $\mathbf{A}' \gets \operatorname{kNN}(\overline{\mathbf{X}}; k)$ via Eq.~(\ref{eq:knn})}
Calculate $\overline{\mathbf{X}}' \gets \operatorname{propagation}(\mathbf{X}, \mathbf{A}'; T, \alpha)$ via Eq.~(\ref{eq:prop})\\
\tcc{Model Training}
Initialize model parameters\\
\For{$e=1,2,\cdots,E$}{
    Calculate $\mathbf{H},\overline{\mathbf{Y}} \gets \operatorname{transformation}(\overline{\mathbf{X}})$ via Eq.~(\ref{eq:trans})\\
    Calculate $\mathbf{H}',\overline{\mathbf{Y}}' \gets \operatorname{transformation}(\overline{\mathbf{X}}')$ via Eq.~(\ref{eq:trans})\\
    Calculate $\mathcal{L}_{ce},\mathcal{L}'_{ce}$ from $\overline{\mathbf{Y}},\overline{\mathbf{Y}}',\mathbf{Y}_L$;\\
    Calculate $\mathbf{Z},\mathbf{Z}'$ from $\mathbf{H},\mathbf{H}'$ by the linear projection layer;\\
    Calculate $\mathbf{p}_1,\cdots,\mathbf{p}_c \gets \operatorname{allocate-proto}(\mathbf{Z},\overline{\mathbf{Y}})$ via Eq.~(\ref{eq:proto})\\
    Calculate $\mathbf{p}'_1,\cdots,\mathbf{p}'_c \gets \operatorname{allocate-proto}(\mathbf{Z}',\overline{\mathbf{Y}})$ via Eq.~(\ref{eq:proto})\\
    Calculate $\mathcal{L}_{cpa}$ from $\mathbf{p}_1,\mathbf{p}'_1,\cdots,\mathbf{p}_c,\mathbf{p}'_c$ and $\tau$ via Eq.(\ref{eq:cpa})\\
    Calcluate $\mathcal{L}$ from $\mathcal{L}_{ce},\mathcal{L}'_{ce},\mathcal{L}_{cpa}$ and $\gamma_1,\gamma_2$ via Eq.(\ref{eq:loss})\\
    Update model parameters via gradient descent\\
}
\vspace{-1mm}
\end{algorithm}

The training algorithm of \ourmethod is summarized in Algo.~\ref{alg:overall}.

\subsection{Complexity Analysis} \label{appendix:complexity}

In this subsection, we discuss the time complexity of \ourmethod at the pre-processing phase and model training phase, respectively. 

During pre-processing, for the original channel and global channel, the complexities of propagation are $\mathcal{O}(mdT)$ and $\mathcal{O}(nkdT)$, respectively. 
The complexity of kNN is $\mathcal{O}(n^2d)$ for full-node kNN and $\mathcal{O}(nbd)$, as discussed in Appendix~\ref{appendix:fast_knn}. To sum up, the time complexities of pre-processing phase is $\mathcal{O}((m+nk)dT+n^2d)$; with scalability extension, the complexity drops to $\mathcal{O}((m+nk)dT+nbd)$.

For the model training phase, we first discuss the complexity of \ourmethod without scalability extension. The complexity of transformation is $\mathcal{O}(nde + nce)$ for the two-layer MLP. For the calculation of $\mathcal{L}_{cpa}$, the complexity of projection and loss computation are $\mathcal{O}(e^2n)$ and $\mathcal{O}(c^2e)$, respectively. The complexity of prototype allocation is a smaller term that can be omitted. 
Hence, the overall time complexity per training epoch is $\mathcal{O}(ne(e+d+c)+c^2e)$. With scalability extension, $n$ can be reduced to $n_L+n_B$, then the complexity becomes $\mathcal{O}(e(n_L+n_B)(e+d+c)+c^2e)$.

\section{Details of Experiments} \label{appendix:exp_detail_full}

\subsection{Datasets} \label{appendix:dset}

\begin{table}[t!]
\centering
\caption{Statistics of datasets.}
\vspace{-2mm}
\resizebox{1\columnwidth}{!}{
\begin{tabular}{l|ccccc}
\toprule
{Dataset} & {\#Nodes} & {\#Edges}  & {\#Classes} & {\#Features} \\ \hline
Cora       & 2,708    & 5,429     & 7       & 1,433  \\ 
CiteSeer   & 3,327    & 4,732     & 6       & 3,703  \\ 
PubMed	   & 19,717   & 44,338    & 3       & 500    \\ 
Amazon Photo & 7,650    & 238,162   & 8       & 745    \\ 
Amazon Computers & 13,752   & 491,722   & 10      & 767   \\ 
CoAuthor CS & 18,333   & 163,788   & 15      & 6,805    \\ 
CoAuthor Physics & 34,493   & 495,924   & 5       & 8,415 \\ 
ogbn-arxiv & 169,343  & 1,166,243 & 40      & 128    \\ 

\bottomrule
\end{tabular}}
\label{tab:dataset}
\vspace{-2mm}
\end{table}

We evaluate our models on eight real-world benchmark datasets for node classification tasks, including Cora, CiteSeer, PubMed, Amazon Photo, Amazon Computers, CoAuthor CS, CoAuthor Physic, and ogbn-arxiv~\cite{sen2008collective, shchur2018pitfalls, hu2020open}. The statistics of datasets are provided in Table~\ref{tab:dataset}. The details are introduced as follows:

\begin{itemize}
    \item \textbf{Cora, CiteSeer} and \textbf{PubMed} \cite{sen2008collective} are three citation networks, where each node is a paper and each edge is a citation relationship. The features are the bag-of-word embeddings of paper context, and labels are the topic of papers. 
    \item \textbf{Amazon Photo} and \textbf{Amazon Computers} \cite{shchur2018pitfalls} are two co-purchase networks from Amazon, where each node is an item, and each edge is a co-purchase relationship between two items. The features include the bag-of-word embeddings of item reviews, and the labels are the category of goods.
    \item \textbf{CoAuthor CS} and \textbf{CoAuthor Physics} \cite{shchur2018pitfalls} are two co-authorship networks from Microsoft Academic Graph. In these graph datasets, each node is an author and each edge is a co-authorship relationship between two authors. The features are the bag-of-words embeddings of the keyword of papers, and the labels are the research directions of authors. 
    \item \textbf{ogbn-arxiv} \cite{hu2020open} is a citation network with Computer Science arXiv papers. In ogbn-arxiv, each node is a paper and each edge is a citation relationship between two papers. 
    The features are the word embeddings of titles/abstracts of papers, and the labels are 40 subject areas of papers. 
\end{itemize}

\subsection{Implementation Details} \label{appendix:exp_detail}

\noindent\textbf{Hyper-parameters. }
We perform small grid search to choose the key hyper-parameters in \ourmethod, where a group of hyper-parameters is searched together. The search space is provided as follows:

\begin{itemize}
    \item Propagation iterations $T$: \{5, 10, 15, 20\}
    \item Restart probability $\alpha$: \{0.01, 0.05, 0.1, 0.2\}
    \item Hidden unit $e$: \{32, 64, 128, 256\} (for datasets excepted ogbn-arxiv);  \{256, 512, 1024\} (for ogbn-arxiv)
    \item Learning rate: \{0.1, 0.05, 0.01, 0.005\}
    \item Number of epochs: \{200, 500, 1000, 2000, 5000\}
    \item Weight decay: \{5e-3, 5e-4\}
    \item Trade-off coefficient $\gamma_1$: \{0.5, 1, 1.5, 2, 3, 5, 8, 10, 15, 20\}
    \item Trade-off coefficient $\gamma_2$: \{0.5, 1, 1.5, 2, 3, 5, 8, 10, 15, 20\}
    \item Number of neighbor in kNN $k$: \{5, 10, 15, 20\}
    \item Metric in kNN: \{`cosine', `minkowski'\}
\end{itemize}

We set the dropout rate to $0.5$, and set the temperature $\tau$ in contrastive loss to $0.3$. For ogbn-arxiv, we set the batch size to $1024$; for the rest datasets, we use all nodes to train the model. 

\noindent\textbf{Computing infrastructures. }
We implement the proposed methods with PyTorch 1.12.1 and PyTorch Geometric 2.1.0. We conduct the experiments on a Linux server with an Intel Xeon E-2288G CPU and two Quadro RTX 6000 GPUs (24GB memory each).